\documentclass[sigconf]{acmart}
\usepackage{balance}
\usepackage{enumitem}
\usepackage{caption}
\usepackage{multirow}
\usepackage{hyperref}

\AtBeginDocument{%
  \providecommand\BibTeX{{%
    \normalfont B\kern-0.5em{\scshape i\kern-0.25em b}\kern-0.8em\TeX}}}

\copyrightyear{2022} 
\acmYear{2022} 
\setcopyright{acmcopyright}\acmConference[MM '22]{Proceedings of the 30th ACM
International Conference on Multimedia}{October 10--14, 2022}{Lisboa, Portugal}
\acmBooktitle{Proceedings of the 30th ACM International Conference on Multimedia
(MM '22), October 10--14, 2022, Lisboa, Portugal}
\acmPrice{15.00}
\acmDOI{10.1145/3503161.3547790}
\acmISBN{978-1-4503-9203-7/22/10}

\acmSubmissionID{236}

\begin{document}

\title{AI Illustrator: Translating Raw Descriptions into Images by Prompt-based Cross-Modal Generation}


\author{Yiyang Ma}
\authornote{This work was done while Yiyang Ma was a research intern at Microsoft Research Asia.}
\affiliation{%
  \institution{Wangxuan Institute of Computer Technology, Peking University}
  \city{Beijing}
  \country{China}
}
\email{myy12769@pku.edu.cn}

\author{Huan Yang}
\affiliation{%
  \institution{Microsoft Research}
  \city{Beijing}
  \country{China}
}
\email{huayan@microsoft.com}

\author{Bei Liu}
\affiliation{%
  \institution{Microsoft Research}
  \city{Beijing}
  \country{China}
}
\email{bei.liu@microsoft.com}

\author{Jianlong Fu}
\affiliation{%
  \institution{Microsoft Research}
  \city{Beijing}
  \country{China}
}
\email{jianf@microsoft.com}

\author{Jiaying Liu}
\authornote{Corresponding author. This work is supported by the National Natural Science Foundation of China under Contract No.62172020.}
\affiliation{%
  \institution{Wangxuan Institute of Computer Technology, Peking University}
  \city{Beijing}
  \country{China}
}
\email{liujiaying@pku.edu.cn}

\renewcommand{\shortauthors}{Yiyang Ma et al.}
\newcommand{\eg}{\textit{e}.\textit{g}.}

\begin{abstract}
AI illustrator aims to automatically design visually appealing images for books to provoke rich thoughts and emotions. 
To achieve this goal, we propose a framework for translating raw descriptions with complex semantics into semantically corresponding images.
The main challenge lies in the complexity of the semantics of raw descriptions, which may be hard to be visualized (\eg, ``gloomy'' or ``Asian''). 
It usually poses challenges for existing methods to handle such descriptions.
To address this issue, we propose a \textbf{P}rompt-based \textbf{C}ross-\textbf{M}odal Generation \textbf{Frame}work (PCM-Frame) to leverage two powerful pre-trained models, including CLIP and StyleGAN.
Our framework consists of two components: a projection module from \textit{Text Embedding}s to \textit{Image Embedding}s based on prompts, and an adapted image generation module built on StyleGAN which takes \textit{Image Embedding}s as inputs and is trained by combined semantic consistency losses. 
To bridge the gap between realistic images and illustration designs, we further adopt a stylization model as post-processing in our framework for better visual effects.
Benefiting from the pre-trained models, our method can handle complex descriptions and does not require external paired data for training.
Furthermore, we have built a benchmark that consists of 200 raw descriptions. 
We conduct a user study to demonstrate our superiority over the competing methods with complicated texts. We release our code at \href{https://github.com/researchmm/AI\_Illustrator}{https://github.com/researchmm/AI\_Illustrator}.
\end{abstract}




%

\maketitle

\begin{figure}[htbp]
  \centering
  \includegraphics[height=5.3cm]{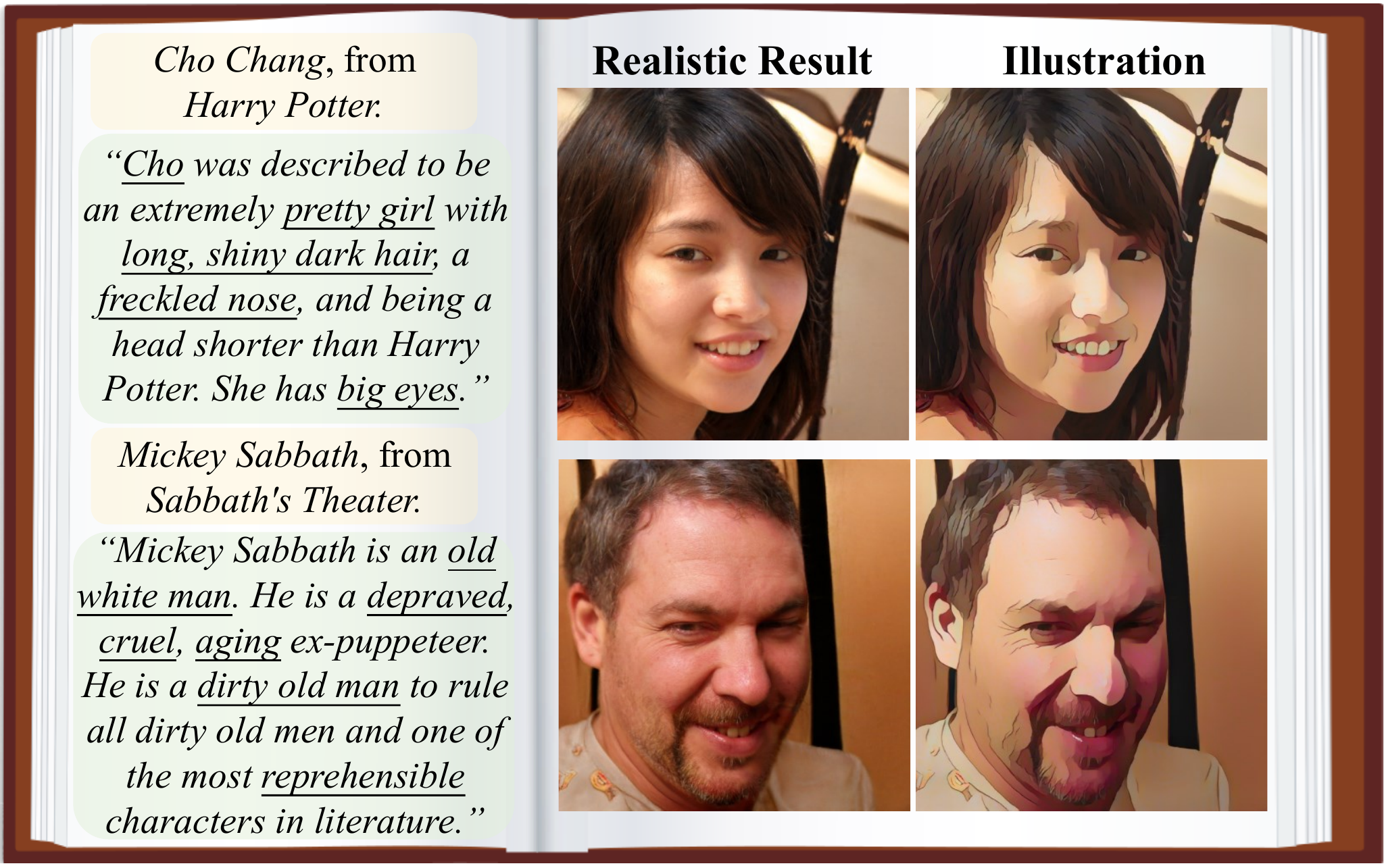}
  \caption{Illustrations generation examples of the proposed AI Illustrator framework. The realistic images are further style transferred as the final illustration design. The \textit{descriptions} are raw descriptions obtained from the Internet or excerpted from books. The major attributes are \underline{underlined}. An interesting fact is that even \textit{Cho Chang} is not explicitly expressed as an Asian girl, our method can infer this from her name and generate an ethnic Chinese girl.}
  \label{fig:teaser}
\end{figure}

\section{Introduction}
\label{sec:Intro}
Characters or scenarios in books often attract readers to image what they really look like. People expect that there could be a method to translate raw descriptions into images in order to help people imagine. The result images of translation must match the descriptions in the aspect of semantics. In the past, this work can only be done by humans, since there are two challenges. The first challenge is the raw descriptions can be long and complicated, and are hard to be visualized precisely. The descriptions not only can be specific, \eg, skin color or hair length, but also can be abstract, \eg, the emotions or personalities. And there may be multiple semantics in one description. The second challenge is that the images should have fine-grained attributes which should be impressive. In this work, we pay attention to dealing with such descriptions with long sentences and complex semantics, and illustrate them with semantic consistency and high quality.

Existing methods \cite{AttnGAN, SD-GAN, MirrorGAN, DM-GAN, ControlGAN, DF-GAN, TediGAN-A, TediGAN-B, StyleCLIP} of text-to-image translation have achieved great results due to the recent cross-modal researches. However, such complex descriptions are difficult for them to handle. \citet{AttnGAN} and \citet{DF-GAN}, use paired data to train, so their performances heavily rely on the quality of datasets and they cannot process those words out of the vocabulary. That means it is nontrivial for them to translate raw descriptions in our task. \citet{TediGAN-B} and \citet{StyleCLIP}, have the ability to deal with complex texts, while they fail in some cases (\eg, text contains rich semantics, as shown in Figure \ref{fig:ComplicatedComparison}) because their optimization methods cannot take full advantage of pre-trained vision-language models. Besides, optimization methods are also time-consuming so it is unpractical to widely utilize these methods. In conclusion, previous methods cannot resolve the two challenges we mention.

In this work, we propose a simple but effective framework named \textbf{P}rompt-based \textbf{C}ross-\textbf{M}odal Projection \textbf{Frame}work (PCM-Frame) to translate raw descriptions into illustrations. Our method can resolve the two challenges above because of the appropriate design with two pre-trained models including StyleGAN and Contrastive Language-Image Pre-training (CLIP). However, how to jointly leverage the two pre-trained models is nontrivial and challenging. That is because there are enormous gaps between the different latent spaces of StyleGAN and CLIP and it is hard to bridge these gaps.

To achieve this goal, we introduce two novel modules in our method. First, a prompt-based projection module which projects \textit{Text Embedding}s to \textit{Image Embedding}s is proposed. Prompt, as usually a method of adapting pre-trained models to downstream tasks, has been extensively utilized in natural language processing (NLP) problems \cite{prompt-survey}. In our work, they are used to connect the two latent spaces of CLIP, representing ``a normal description'' and ``a normal image'' in order to be on behalf of the whole latent space. CLIP encodes texts to \textit{Text Latent Space} and images to \textit{Image Latent Space}. Semantically aligned image-text pairs will be encoded to embedding pairs which have high cosine similarity, and vice versa. In this module, we take a specific pair of \textit{Text Embedding} and \textit{Image Embedding} as ``prompts'' and migrate the \textit{input Text Embedding} which is extracted from the input text description to \textit{input Image Embedding} by the connection of \textit{prompt embedding}s. Second, we propose a module which projects \textit{Image Embedding}s to \textit{StyleGAN Embedding}s and generates images from them by StyleGAN. StyleGAN \cite{StyleGANv1, StyleGANv2}, as one of the most notable Generative Adversarial Network (GAN) \cite{GAN} frameworks, helps us to generate images with high quality. This module contains a network trained on paired data which are randomly sampled. We randomly sample \textit{StyleGAN Embedding}s, generate images from them by pre-trained StyleGAN and extract \textit{Image Embedding}s from them by CLIP. The training of this projection module does not require any external data. In order to train the network with semantics maintained, we further design combined semantic consistency losses for the training of this network to ensure we can generate semantically accurate images by StyleGAN. At last, we use a style transfer method \cite{Cartoonize} to cartoonize them so that these images can be used as illustrations for books. Two results of our method are provided in Figure \ref{fig:teaser}. 

Besides, we build a benchmark consisting of 200 raw descriptions of characters which are challenging to visualize. We evaluate the performance of our method on some of the descriptions and conduct a user study on these translation results. In summary, this work has the following contributions:

\begin{itemize}[nosep, left=0.5cm]
    \item We propose a framework that can translate complicated raw text descriptions into illustrations with high semantic consistency, quality, and fidelity.
    \item We propose a novel prompt-based projection from \textit{Text Embedding}s to \textit{Image Embedding}s. We also propose a loss function which helps our framework keep the semantic consistency and the training process doesn't require any external paired data. 
    \item Experiments demonstrate the superiority of our method. The user study shows that more than 89\% of 2200 votes from 22 subjects prefer our method to other state-of-the-art methods.
\end{itemize}

\begin{figure*}
    \centering
    \includegraphics[width=\linewidth]{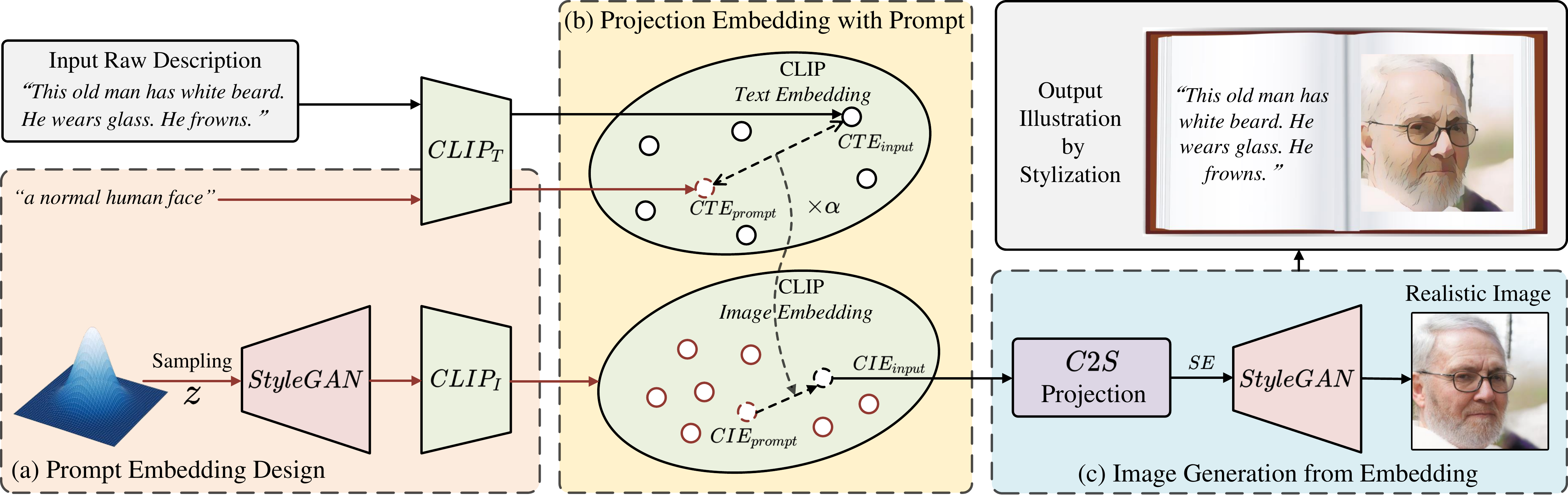}
    \caption{The entire framework of PCM-Frame. $CLIP_{T}$ and $CLIP_{I}$ denote the text encoder and image encoder of CLIP. Beforehand, in (a), we design $CTE_{prompt}$ and $CIE_{prompt}$ which are expressed as red dotted circles and used in (b). The $CTE_{prompt}$ is extracted by CLIP from a certain sentence while the $CIE_{prompt}$ is obtained from a big set of $CIE$s (expressed as red circles). At inference time, the input raw text description is encoded to $CTE_{input}$ by CLIP and projected to the corresponding $CIE_{input}$ in (b). Then, the $CIE_{input}$ is projected to the corresponding $SE$ from which we can generate a semantically aligned realistic image by StyleGAN in (c). At last, the generated realistic image can be further style transferred as the final illustration design. The specific architecture of $C2S$ projection network is shown in Figure \ref{fig:C2S}.}
    \label{fig:WholePipeline}
\end{figure*}

\section{Related work}
\label{sec:RelatedWork}

\subsection{Text-to-Image Semantic Alignment}
\label{subsec:T2I_AlignmentRelatedWork}

In order to translate a text into an image, checking whether the text and the image are semantically aligned is necessary. 

In the early years, works like \cite{GAN-INT-CLS, StackGAN, SISGAN, StackGAN++, HD-GAN, AttnGAN, Obj-GAN} train a pair of text encoder and image encoder separately to semantically align them. \citet{GAN-INT-CLS} train two simple networks to extract embeddings from texts and images respectively. For those pairs of text and image which relatively semantically aligned, the embedding pairs will have a higher dot product and vice versa. \citet{AttnGAN} propose a method with an attention mechanism called Deep Attentional Multimodal Similarity Model (DAMSM). \citet{DM-GAN, DF-GAN, ManiGAN, Lightweight-Manipulation} also use this model. All the methods above need to be trained for a certain translation task. So, their performance heavily relies on the quality of the datasets and they cannot encode the words out of the vocabulary of the datasets.

The methods of extracting fine grained level features of images \cite{LookCloser, LearningMultiAttention} are also inspiring. Following the success of BERT \cite{BERT} in language tasks, recent works typically use transformers \cite{Attention} as baseline models. A recent model, based on Contrastive Language-Image Pre-training (CLIP) \cite{CLIP}, builds two encoders of transformer for a wide range of images and texts. CLIP is trained on 400 million text-image pairs which are collected from a variety of publicly available sources on the Internet. There are two latent spaces, one for texts and another for images. In our work, we take CLIP as our text-to-image alignment checking module. 

\subsection{Text-to-Image Translation}
\label{subsec:T2I_TranslationRelatedWork}

Due to the great development of GANs in many fields \cite{AOT-GAN, TTSR, ImproveVisualQuality, PEN-Net, STTN, MetroGAN}, most existing text-to-image translation methods are GAN-based. They can be roughly divided into two categories.

The first category doesn't use pre-trained generation models, \eg, StyleGAN \cite{StyleGANv1, StyleGANv2}. These methods train an image generator themselves. The pioneering work of \citet{GAN-INT-CLS} approaches text-to-image translation by training a conditional GAN \cite{cGAN} with text embeddings extracted from a pre-trained text encoder. \citet{AttnGAN} introduce an attention module between images and texts.
Following the attention module which is proposed by \citet{AttnGAN}, \citet{DM-GAN} introduce a memory writing gate and \citet{DF-GAN} propose a backbone that generates images directly by Wasserstein distance. \citet{DALL-E} build a large model with over 12-billion parameters and show a great diversity of text-to-image translation. 

The second category uses existing generation models so that the quality of their generated images is better and the training process can be shorter. But because of the domain limitation of the existing generation models, the images they generate are limited in certain domains. \citet{TediGAN-A} map input text to StyleGAN latent space, while \citet{TediGAN-B} use cosine similarity of text and image embeddings encoded by CLIP as a loss function to optimize an embedding in StyleGAN latent space. Due to the usage of CLIP, \citet{TediGAN-B} can process texts with more complex semantics. But its performance is random and visually unpleasant. \citet{StyleCLIP} propose three methods to manipulate an existing image. The first method of latent optimization they propose can be used as an image generation method by giving an initial image. \citet{StyleGAN-nada} transfer images to new text-guided domains by fine-tuning StyleGAN. We use StyleGAN2 \cite{StyleGANv2} in our work.

\subsection{Prompt Method in NLP Domain}
\label{subsec:PromptRelatedWork}

Prompt is a recently proposed method to ``re-formulate'' downstream tasks so that these tasks can be resolved by pre-trained models \cite{prompt-survey}. It can be regarded as an ``intermediate''. It is first introduced to solve NLP problems. \citet{prompt-survey} introduce such an example: when recognizing the emotion of a social media post, ``I missed the bus today.'', we may continue with a prompt ``I felt so \_\_\_.'', and ask the language model (LM) to fill the blank with an emotion-bearing word instead of giving the LM only the sentence ``I missed the bus today.'' which do not have a precise task. In this way, the downstream task is re-formulated so that a pre-trained LM can handle. Without such ``prompt''s, if we want to leverage pre-trained models on a downstream task, the models have to be finetuned on the corresponding data, while finetuning is time-consuming and the performances heavily rely on the quality of the dataset. 

In NLP domain, there have been many automatic methods \cite{Autoprompt, PromptProgramming, OptimizePrompt} of designing prompts for certain tasks instead of manually specifying \cite{PromptBasic1, GPTUnderstands}. We refer the reader to the survey \cite{prompt-survey} for extensive exposition and discussion on prompt. The methods of prompt have been widely used in NLP domain. No attempt of applying the prompt method in Cross-Modal problems or Computer Vision problems so far. In our work, we leverage the idea of prompt and propose a specific pair of embeddings as ``prompt''s to help us bridge the two latent spaces of CLIP. Our method is a novel understanding of prompts which is different from the methods proposed in NLP.

\section{Prompt-based Cross-Modal Generation Framework}
\label{sec:Approach}

In this section, we will introduce our AI Illustrator framework that can translate raw text descriptions into vivid illustrations. Our framework consists of two major modules which is shown in Figure \ref{fig:WholePipeline}. The first module in (b) projects \textit{Text Embedding}s (abbreviated as $CTE$ and $C$ denotes CLIP) to \textit{Image Embedding}s (abbreviated as $CIE$ and $C$ denotes CLIP) with \textit{prompt embedding}s and the second module in (c) generates images from the projected $CIE$s. In particular, we encode the input text to $CTE_{input}$ using CLIP and project it to $CIE_{input}$ via the first module with a pair of \textit{prompt embedding}s. Then, via the second module, the $CIE_{input}$ is projected to the corresponding \textit{StyleGAN $\mathcal{Z}$ Embedding} (abbreviated as $SE$, and we use StyleGAN2 in our work) and we generate an image from it which is semantically aligned to the original text by pre-trained StyleGAN. To bridge the gap between realistic images and illustrations, we style transfer the generated images at last. We do not build a direct projection from input text to $SE$ because there are few paired data. So, we take $CIE$ as a transition to connect the texts and $SE$s.

\subsection{Prompt Embedding Design}
\label{subsec:Prompt}

As we point out in Section \ref{sec:Intro}, $CTE_{prompt}$ and $CIE_{prompt}$ are used to bridge the text and image latent spaces of CLIP and these prompts should represent ``a normal description'' and ``a normal image'' respectively. Otherwise, there may be a big distance between the prompt pair and the target pair which may lead to failure. So, it is important to assign an appropriate pair as prompts. Our prompt design is shown in the Figure \ref{fig:WholePipeline} (a). The specific method of using these prompts will be further explained in Section \ref{subsec:Projection1}.

To make sure that the prompt can represent all data, the \textit{prompt embedding} should be extracted from a big enough set of data. We assume that the \textit{prompt embedding} should have the largest average cosine similarity to all the embeddings in the set. Their lengths are all normalized because only their orientations contain semantics. Taking $\boldsymbol{y}$ to denote the \textit{prompt embedding} and $\boldsymbol{x_{i}}$ to denote the $i$-th the embeddings in the set, the problem can be formulated as
\begin{equation}
\label{equ:3}
    \max \limits_{\boldsymbol{y}} z = \frac{1}{n} \sum_{i=1}^{n} \frac{\boldsymbol{y} \cdot \boldsymbol{x_{i}}}{|\boldsymbol{y}| \cdot |\boldsymbol{x_{i}}|},
\end{equation}
\begin{equation}
\label{equ:4}
    s.t. |\boldsymbol{y}| = 1,
\end{equation}
where $\cdot$ denotes the dot product of vectors, $n$ denotes the number of data in the set and $z$ denotes the average cosine similarity between the \textit{prompt embedding} and all other embeddings.
Note that this is a non-linear programming problem which is hard to resolve directly. To address the above issue and obtain \textit{prompt embedding}s, we propose to find the physical meanings of the equations which will help us to solve this problem. 

Because the lengths of all of the embeddings are all normalized, Equation \ref{equ:3} can be simplified to 
\begin{equation}
\label{equ:5}
    \max \limits_{\boldsymbol{y}} z = \frac{1}{n} \sum_{i=1}^{n} \boldsymbol{y} \cdot \boldsymbol{x_{i}}.
\end{equation}

From Equation \ref{equ:5}, by using associative law and commutative law of addition and multiplication, we can get 
\begin{equation}
    \max \limits_{\boldsymbol{y}} z = \boldsymbol{y} \cdot \frac{1}{n} \sum_{i=1}^{n} \boldsymbol{x_{i}}.
\end{equation}

The above equation represents a hyperplane and $z$ is the constant parameter. The farther the hyperplane is from the origin point, the larger the absolute value of $z$ is. And as the feasible region of this problem is a hypersphere with symmetry which is shown in Equation \ref{equ:4}, we can move this hyperplane as far as possible if we want to maximize $z$. It is easy to see that $z$ will be the largest when the hyperplane and the hypersphere are tangent and $\boldsymbol{y}$ is the unit normal vector of the hyperplane at this moment. Through analytic geometry, the unit normal vector of the hyperplane is a vector as shown below:
\begin{equation}
    \boldsymbol{y'} = \frac{1}{n} \sum_{i=1}^{n}\boldsymbol{x_{i}},\  \boldsymbol{y} = \frac{\boldsymbol{y'}}{|\boldsymbol{y'}|}.
\end{equation}

It can be seen that the vector $\boldsymbol{y'}$ is the arithmetic average of all embeddings with unit length. 

Specifically, for the image prompt, we can build a big set of images by randomly sampling in $\mathcal{Z}$ space of StyleGAN and generating images from such latent embeddings.
For the text prompt, because it's hard to get a big set of descriptive texts with a large semantic range and the text itself is a carrier of semantics, we can manually specify a certain sentence like the methods mentioned in \cite{prompt-survey}. For example, for text-to-human face translation, we can simply set sentence ``A normal human face.'' and extract its $CTE$ as $CTE_{prompt}$. This choice is further discussed in our ablation Section \ref{subsec:Ablations}. But, if there is a text set which has high enough quality, our method can be used to obtain a better $CTE_{prompt}$. 

\subsection{Embedding Projection with Prompt}
\label{subsec:Projection1}

In this section, we encode the input text to $CTE_{input}$ and manage to obtain the corresponding $CIE_{input}$ via the first module shown Figure \ref{fig:WholePipeline} (b) with the prompts we get in last section. The prompts can be regarded as a pair of ``intermediate''s. To be more specific, the $CTE_{prompt}$ is subtracted from the input $CTE_{input}$, then this difference is added to the $CIE_{prompt}$. The result is the semantically corresponding $CIE_{input}$ of the input text. This module is shown in Figure \ref{fig:WholePipeline} (b) and the projection can be formulated as
\begin{equation}
    CIE_{input} = CIE_{prompt} + (CTE_{input} - CTE_{prompt}).
    \label{equ:6}
\end{equation}

From the perspective of prompt, the summation can be regarded as ``re-formulation'' as \citet{prompt-survey} propose. And in contrast, we propose that the subtraction can be regarded as ``de-formulation''. This projection makes our next module of projection between $CIE$s and $SE$s can process $CTE_{input}$ without fine-tuning on $CTE$-$SE$ pairs which are not easy to obtain.
The validity of the linear operations in Equation \ref{equ:6} is supported by the character of CLIP. Further discussions are provided in supplementary materials.


The inference time of this module only uses linear operations among $CTE_{input}$, $CTE_{prompt}$ and $CIE_{prompt}$ to get $CIE_{input}$. That means this module can run quite efficient and stably.

In our work, the difference between $CTE_{input}$ and $CTE_{prompt}$ will be multiplied by a constant to control the distinctiveness. So, the actually used method is 
\begin{equation}
    CIE_{input} = CIE_{prompt} + \alpha \cdot (CTE_{input} - CTE_{prompt}),
\end{equation}
where $\alpha$ is the constant. It can vary in the range of $1$ to $2$ and is set to $1.75$ which is capable for most descriptions empirically.

\begin{figure}
    \centering
    \includegraphics[width=\linewidth]{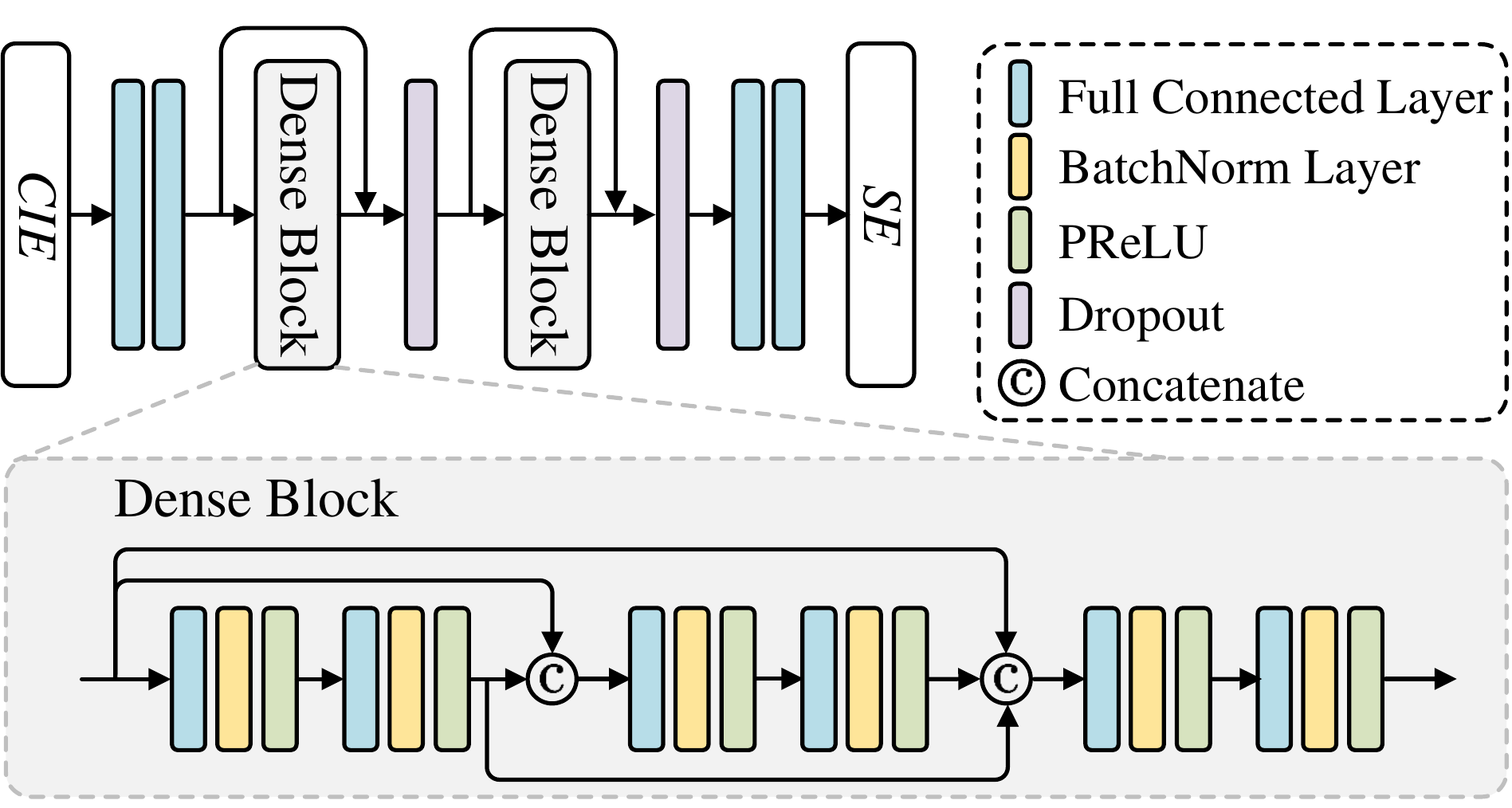}
    \caption{The architecture of CLIP-to-StyleGAN ($C2S$) projection network.}
    \label{fig:C2S}
\end{figure}

\subsection{Image Generation from Embedding}
\label{subsec:Projection2}

After getting the $CIE_{input}$, we manage to generate the corresponding image from it. In the module shown in Figure \ref{fig:WholePipeline} (c), we map the $CIE_{input}$ we get above to the semantically aligned $SE$ from which we can generate the image we expect. We take a combination of fully connected layers with dense connection as this projection and this network is named as CLIP-to-StyleGAN ($C2S$) projection network. The architecture of this network is shown in Figure \ref{fig:C2S}. To train this network, we propose to generate $CIE$-$SE$ pairs. As the distribution of $\mathcal{Z}$ space of StyleGAN is standard normal distribution determinately, we can randomly sample $SE$s in $\mathcal{Z}$ space and generate images from these $SE$s by pre-trained StyleGAN. Thus, we can extract $CIE$s from these images by CLIP. In this way, we can get theoretically infinite $CIE$-$SE$ pairs to train our network. 

The key to this projection is to keep the semantics of $CIE_{input}$, so the loss of this network should be designed purposefully. Besides, we should also guarantee that the $SE$s are within $\mathcal{Z}$ space so that we can generate images by StyleGAN. In order to optimize the above network, we propose the following combined loss functions.

In order to keep the semantics of $CIE_{input}$, we can use CLIP to check the semantic alignment of the generated image and $CIE_{input}$. In particular, we extract the $CIE$ from the generated image as $CIE_{rebuilt}$ and design a loss function called $\mathcal{L}_{sem\_cons}$ to minimize the cosine distance between $CIE_{rebuilt}$ and $CIE_{input}$. The subscript of the loss function is short for \textit{Reconstructing Semantics Consistency}. Taking $G$ to denote pre-trained StyleGAN and $CLIP_I$ to denote the image encoder of CLIP, this loss is calculated by
\begin{equation}
    \mathcal{L}_{sem\_cons} = CosDis(CIE_{input}, CLIP_{I}(G(SE_{pred}))).
\end{equation}

For the basic constraint of the network, we use a l1 loss function called $\mathcal{L}_{l1}$ between $SE_{pred}$ and $SE_{true}$. It is calculated by
\begin{equation}
    \mathcal{L}_{l1} = ||SE_{pred} - SE_{true}||_{1}.
\end{equation}

Besides, the network should make sure that the predicted $SE_{pred}$ is within the $\mathcal{Z}$ space of StyleGAN. Otherwise, StyleGAN cannot generate images from the $SE$s out of the latent space. Because the distribution of $\mathcal{Z}$ space is clearly a standard normal distribution, all the $SE$s should have mean values of 0 and standard deviations of 1. This character helps us to design a regularization loss called $\mathcal{L}_{reg}$ to ensure the generated $SE$s are all within $\mathcal{Z}$ space easily and this is the reason why we use $\mathcal{Z}$ space. This loss is calculated by
\begin{equation}
    \mathcal{L}_{reg} = ||mean(SE_{pred})||_{1} + ||std(SE_{pred}) - 1||_{1}.
\end{equation}

To sum up, the total loss of our network is shown below.
\begin{equation}
    \mathcal{L} = \lambda_{sem\_cons} \cdot \mathcal{L}_{sem\_cons} + \lambda_{l1} \cdot \mathcal{L}_{l1} +  \lambda_{reg} \cdot \mathcal{L}_{reg},
\end{equation}
where $\lambda_{sem\_cons}$, $\lambda_{l1}$ and $\lambda_{reg}$ are the corresponding weights of all the losses. There values are 1.0, 0.3, 0.3 respectively.

After getting the $SE$, we can simply generate the corresponding image from it by StyleGAN. But there is still a gap between the generated images and illustrations because most illustrations are more abstract than realistic images. In order to bridge this gap, we adopt an existing stylization method \cite{Cartoonize} to cartoonize the images at last and get the final design of illustration.

\section{Experimental results}
\label{sec:Experiments}

In this section, we show a lot of experimental results to demonstrate the superiority of our framework and evaluate the effectiveness of the modules we propose.
The implementation details to ensure the reproducibility are provided in the supplementary materials.

\begin{figure}
    \centering
    \includegraphics[height=9.3cm]{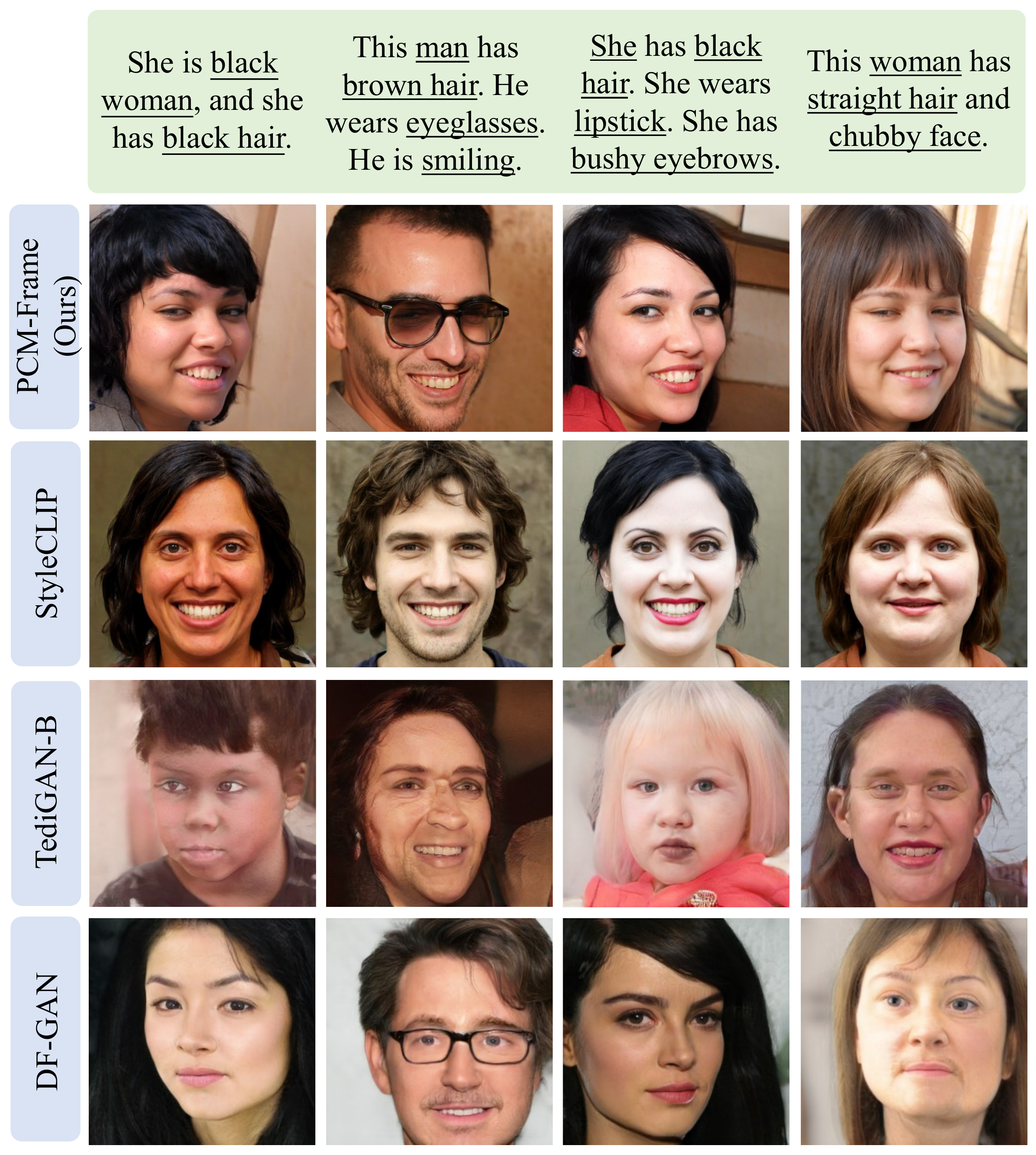}
    \caption{Translation results for human faces on descriptions with simple words by multiple methods. The major attributes are \underline{underlined}.}
    \label{fig:SimpleComparison}
\end{figure}

\subsection{Baseline Methods}
\label{subsec:Baselines}

To demonstrate the superiority of the method we propose, we compare our translation results with state-of-the-art methods including DF-GAN (CVPR 2022) \cite{DF-GAN}, TediGAN-B (arXiv 2021) \cite{TediGAN-B} and StyleCLIP (ICCV 2021) \cite{StyleCLIP}. StyleCLIP, as a work aiming at image manipulation, can also be used for text-to-image translation with the proposed first technique of latent optimization by assigning an origin latent. TediGAN-B and StyleCLIP can process complicated texts due to the usage of CLIP, while DF-GAN cannot.
For the task on human faces, we retrained DF-GAN on the Multi-Modal CelebA-HQ dataset \citet{TediGAN-A}. Other previous methods which need to be retrained \eg, DM-GAN (CVPR 2019) \cite{DM-GAN}, SD-GAN (CVPR 2019) \cite{SD-GAN} and AttnGAN (CVPR 2018) \cite{AttnGAN} get approximate results to DF-GAN and they also cannot process the words out of the vocabulary of the dataset, so we only compare with DF-GAN.

\begin{figure}
    \centering
    \includegraphics[height=8.42cm]{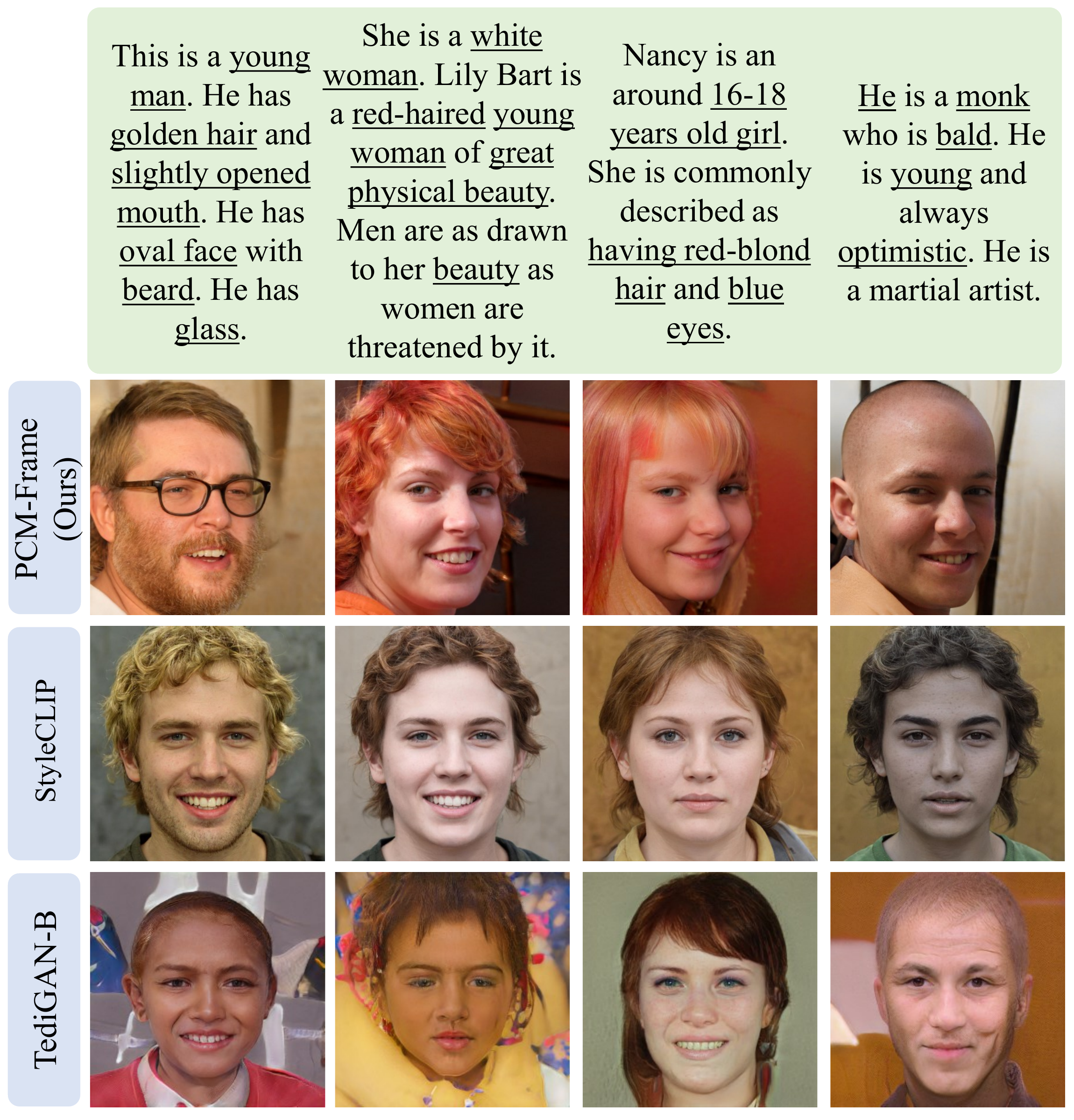}
    \caption{Translation results for human faces on descriptions with complicated words by multiple methods. The major attributes are \underline{underlined}.}
    \label{fig:ComplicatedComparison}
\end{figure}

\begin{figure}
    \centering
    \includegraphics[height=3.45cm]{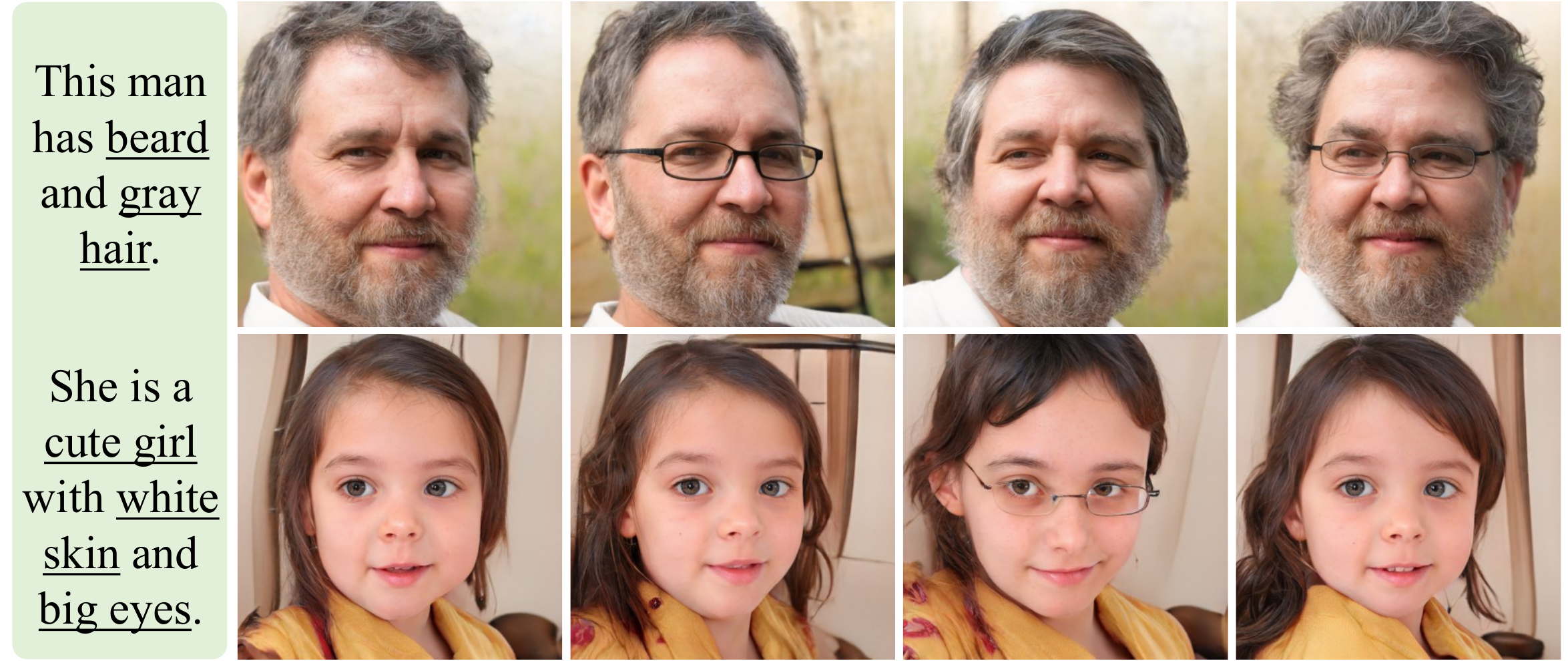}
    \caption{Different results from one description by applying random factors. The major attributes are \underline{underlined}.}
    \label{fig:Diversity}
\end{figure}

\begin{figure*}
    \centering
    \includegraphics[height=9.9cm]{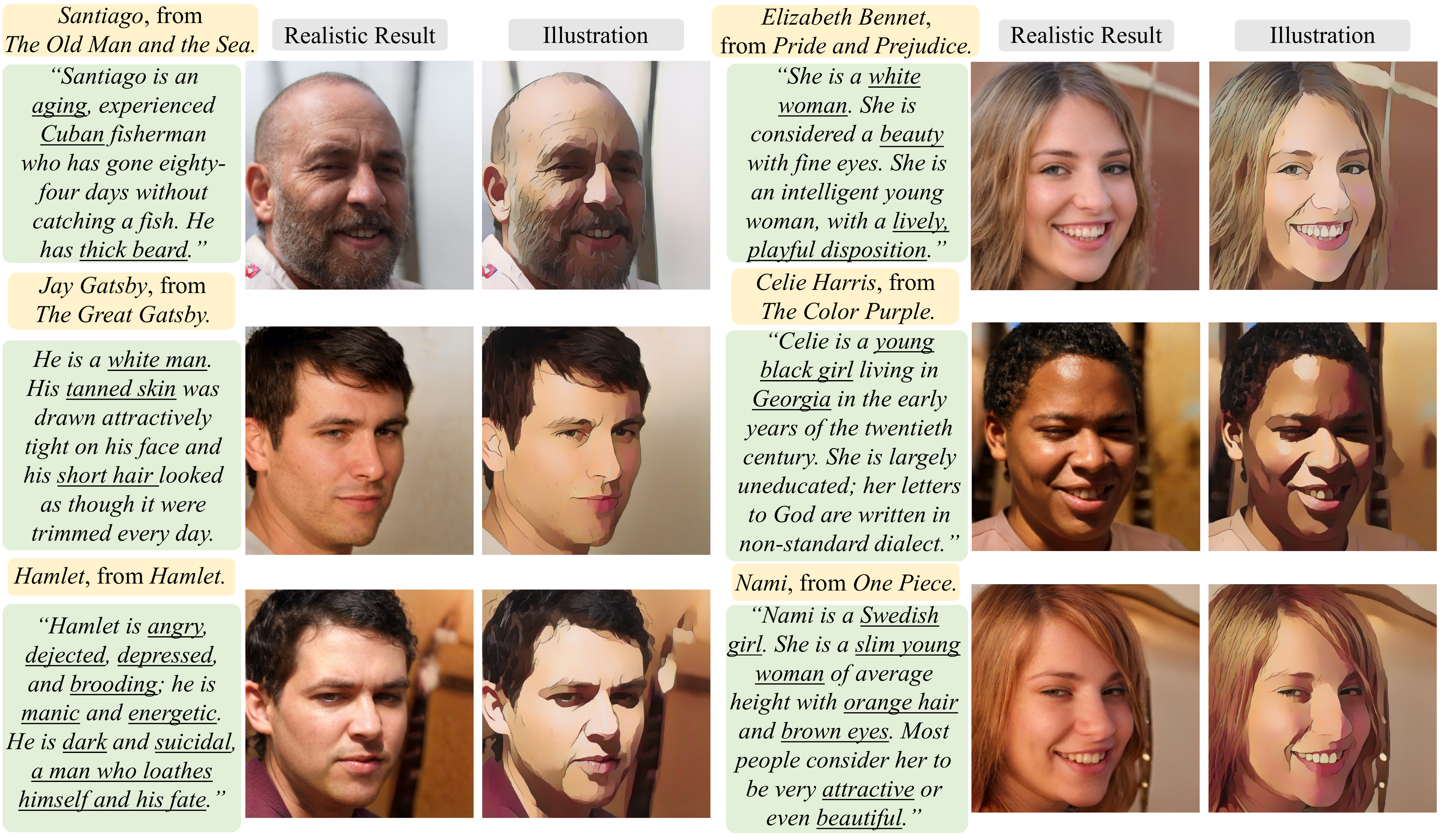}
    \caption{Translation results for the face of famous characters on raw descriptions. The images of the left ones of each pair are our realistic image results while the right ones are style transferred illustrations as the final design. The \textit{descriptions} are all raw descriptions which are obtained from the Internet or excerpted from books. The major attributes are \underline{underlined}.}
    \label{fig:MainResults}
\end{figure*}

\begin{figure*}
    \centering
    \includegraphics[height=9.0cm]{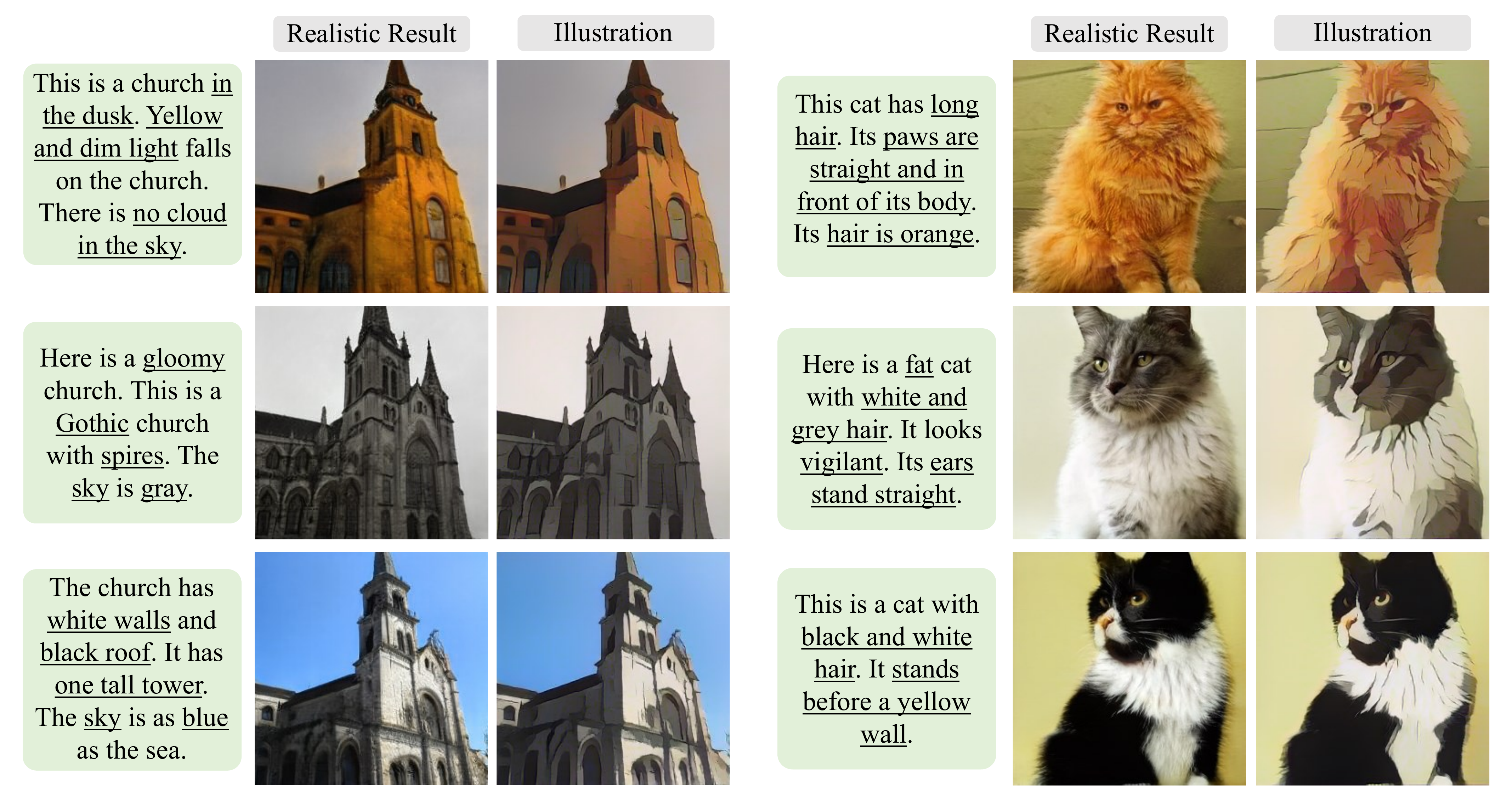}
    \caption{Translation results for non-face descriptions. The images of the left ones of each pair are our realistic image results while the right ones are style transferred illustrations as the final design. The major attributes are \underline{underlined}.}
    \label{fig:MainResults_Church&Cat}
\end{figure*}

\subsection{Comparisons to State-of-the-Art Methods}
\label{subsec:Comparisons}

\textbf{Qualitative Comparison.} First, we qualitatively compare our results with the results of other methods. To demonstrate the ability to translate the input text description to fine-grained images without the interference of style transfer, all the results of this section are not style transferred.
Because DF-GAN can only process the words within the vocabulary of training, we first translate sentences with simple words which appear in the Multi-Modal CelebA-HQ dataset in order to compare with DF-GAN fairly. The results are shown in Figure \ref{fig:SimpleComparison}. It can be seen that our method performs much better than other works. We translate all the semantics of the input text and the images are visually pleasing, while the results of other methods cannot guarantee, \eg, we translate the description of ``black woman'' and the description of ``eyeglasses'' successfully while other methods fail. Then we show translation results on relatively complex texts in Figure \ref{fig:ComplicatedComparison}. There are many descriptions which are hard to be visualized or do not have direct relations to the human face, \eg, ``monk'' and ``16-18 years old''. We translate those indirect descriptions into details in the result images and exclude irrelevant semantics while other methods like StyleCLIP fail. 


We show the diversity of our results in Figure \ref{fig:Diversity} because one description may match multiple images. We translate one text into diverse images by style mixing a random $SE$ in the first few layers of StyleGAN. This is a special mechanism supported by StyleGAN. In order to make every result shown in this paper reproducible, all other results are generated without any random style mixing.

\begin{table}[]
    \centering
    \begin{tabular}{p{0.4cm}p{1.87cm}p{0.6cm}p{1.1cm}p{1.1cm}p{1.15cm}}
        \toprule
        \ & \ & Ours & StyleCLIP \cite{StyleCLIP} & TediGAN-B \cite{TediGAN-B} & DF-GAN \cite{DF-GAN}
        \\
        \midrule
        \multirow{2}{*} {Com.} & Acc. Prefer.(\%) & \textbf{90.6} & 5.1 & 4.3 & -
        \\
        & Real. Prefer.(\%) & \textbf{78.3} & 20.8 & 0.9 & -
        \\
        \midrule
        \multirow{2}{*} {Sim.} & Acc. Prefer.(\%) & \textbf{83.8} & 9.5 & 1.9 & 4.8
        \\
        & Real. Prefer.(\%) & \textbf{75.5} & 23.8 & 0.2 & 0.5
        \\
        \bottomrule
    \end{tabular}
    \caption{The user study for translation results on text descriptions. ``Com.'' denotes ``with Complicated Words'' and ``Sim.'' denotes ``with Simple Words''. The best numbers are \textbf{bold}. For descriptions with complicated words, there are 1760 votes. For descriptions with simple words, there are 440 votes.}
    \label{tab:UserStudy}
\end{table}

We also conduct a user study on the results for human faces of both simple and complicated texts. The users are asked to judge which one is the most photo-realistic (abbreviated as Real. Prefer.) and the most semantically aligned (abbreviated as Acc. Prefer.) to the given text. The cases with simple words contain 20 text-image pairs and cases with complicated words contain 80 text-image pairs. A total of 22 subjects participates in this user study and 2200 votes are collected. The results are shown in Table \ref{tab:UserStudy}. 

\begin{table}[]
    \centering
    \begin{tabular}{p{1cm}p{1.15cm}p{1.15cm}p{1.15cm}p{1.15cm}}
        \toprule
         & Ours & StyleCLIP \cite{StyleCLIP} & TediGAN-B \cite{TediGAN-B} & DF-GAN \cite{DF-GAN}
        \\
        \midrule
        IS $\uparrow$ & \textbf{3.229} & 1.323 & 3.191 & 2.503
        \\
        \bottomrule
    \end{tabular}
    \caption{Inception score comparison of generated results from different methods. $\uparrow$ means the higher the better.}
    \label{tab:IS_comparison}
\end{table}

\noindent\textbf{Quantitative Comparison.} We evaluate inception score (IS) \cite{InceptionScore} to compare the diversity and quality of the generated images in Table \ref{tab:IS_comparison}. All the results are calculated from 100 samples except DF-GAN. The result of DF-GAN is calculated from 20 samples.

\subsection{Illustration Results of Raw Descriptions}
\label{subsec:TranslationResults}

The final goal of our work is to automatically generate illustrations from raw descriptions. In order to demonstrate the ability of our framework, we show more translation results of book characters in Figure \ref{fig:MainResults} and Figure \ref{fig:teaser}. At the same time, we offer the style transferred images of them. The style transfer method is provided by \citet{Cartoonize}. These transferred images can be directly used as illustrations for books. Some of the descriptions are excerpted from books and some of them are obtained from the Internet. These descriptions contain very complex semantics. It's clear that our method successfully translates the underlined attributes, whether they're specific or abstract. And those less relevant parts of texts without underlines do not have negative effects on our results. Besides, in Figure \ref{fig:teaser}, there's an interesting fact that our method translates the text description of \textit{Cho Chang} into an Asian girl because our method infers this from her name. In order to demonstrate our capability of translating general descriptions, we also show similar illustration results on churches and cats in Figure \ref{fig:MainResults_Church&Cat}.

\begin{figure}
    \centering
    \includegraphics[height=2.4cm]{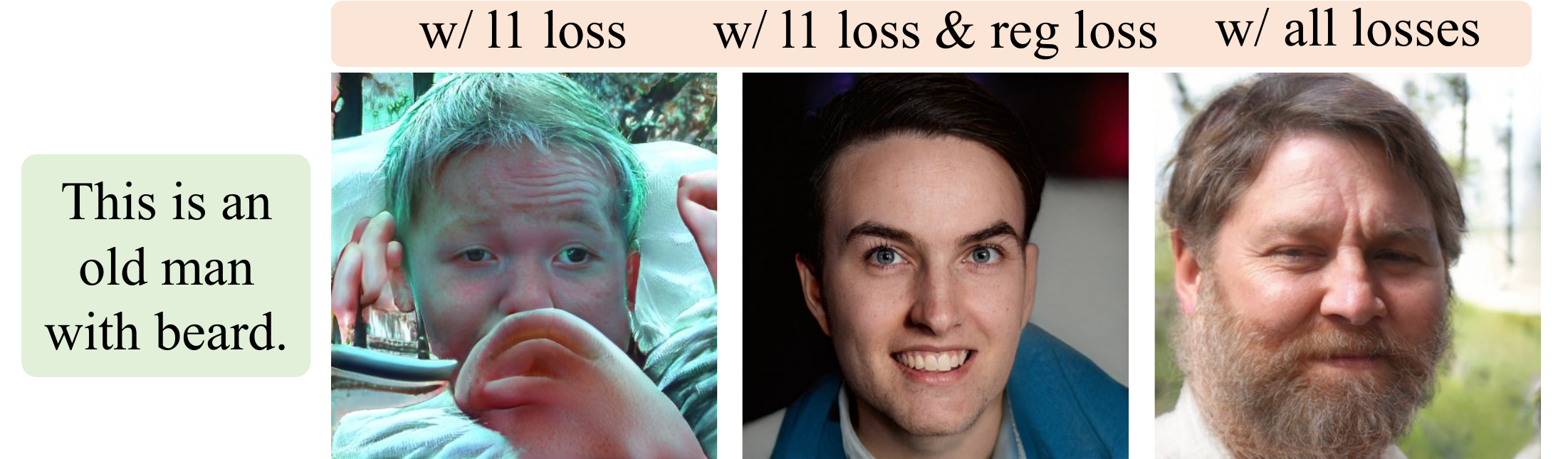}
    \caption{The ablation study on the loss functions we propose. $\mathcal{L}_{reg}$ helps our framework generate images with fidelity of human faces because it guarantees the projected $SE$ is within $\mathcal{Z}$ space. $\mathcal{L}_{sem\_cons}$ keeps the semantics of the input text.}
    \label{fig:LossAblation}
\end{figure}

\begin{figure}
    \centering
    \includegraphics[height=4.8cm]{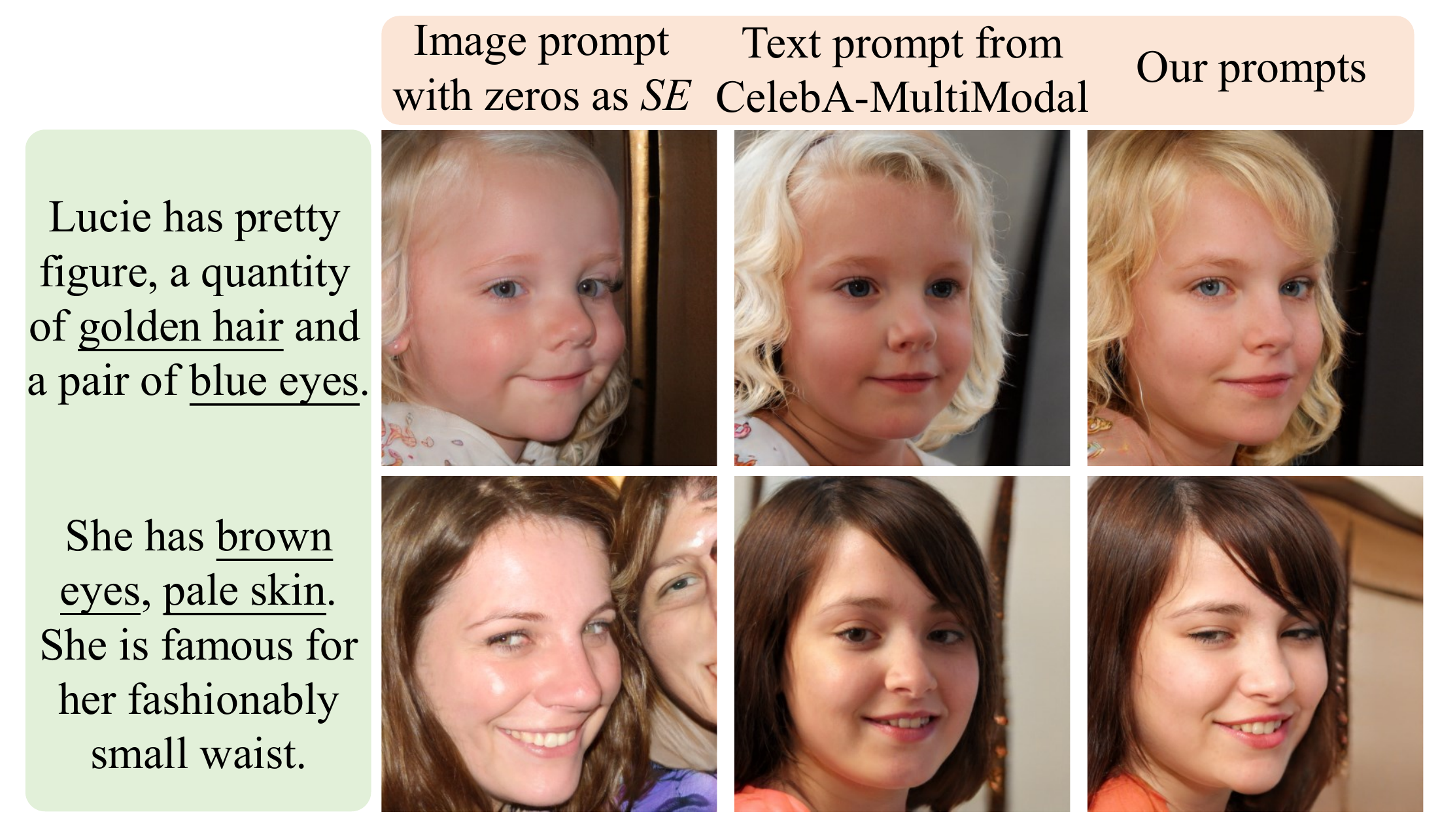}
    \caption{The ablation study on the prompt design. The first column uses the $CIE$ of the image generated from $SE$ which consists of only zeros. The second column uses the average $CTE$ of all the captions from Multi-Modal CelebA-HQ.}
    \label{fig:PromptAblation}
\end{figure}

\subsection{Ablation Study}
\label{subsec:Ablations}

There are two main factors that have effects on the quality of our work, the method of getting prompt and the loss functions of the $C2S$ projection network. We demonstrate their effectiveness by giving an ablation study. In Figure \ref{fig:LossAblation}, we show the results without the proposed losses. In Figure \ref{fig:PromptAblation}, we show ablation study on prompt design. As we discuss in Section \ref{subsec:Prompt}, prompts should represent a normal semantic, we compare our design to several other designs which may contain the semantic of ``normal''. For the image prompt, ours is obtained from a big set of images, so it deserves a better performance. But for the text prompt, the prompt obtained from Multi-Modal CelebA-HQ performs worse than the manually specified text prompt by us, \eg, the corresponding faces look younger than we expect. That's because the quality of this dataset is not high enough. In this dataset, the captions of kids will be indicated as ``young'' while the captions of adults will not be indicated as ``grown''. So the average of these captions will bias to young kids.

Besides, in Table \ref{tab:NetworkAblation}, we provide the comparison of performance between the $C2S$ projection network architecture we propose and a simple MLP with the same number, 54, of fully connected layers. We use the average cosine distance between $CIE$s of generated images and original images to prove the ability of keeping semantically alignment, Fr\'{e}chet Inception Distance (FID) \cite{TTUR} and IS \cite{InceptionScore} to prove the quality and diversity of generated images. The FID is calculated between 100 generated samples and 2000 random samples from FFHQ \cite{StyleGANv1} and the IS is calculated from 100 generated samples.

\begin{table}
    \centering
    \begin{tabular}{p{2cm}p{1.8cm}p{1.8cm}}
        \toprule
         & Ours & MLP
        \\
        \midrule
        $CIE$ Distance $\downarrow$ & \textbf{0.1037} & 0.2852
        \\
        FID $\downarrow$ & \textbf{112.91} & 119.83
        \\
        IS $\uparrow$ & \textbf{3.229} & 3.133
        \\
        \bottomrule
    \end{tabular}
    \caption{The ablation study on network architecture. We compare our $C2S$ projection network and a simple MLP. $\uparrow$ means the higher the better and vice versa.}
    \label{tab:NetworkAblation}
\end{table}

\begin{figure}
    \centering
    \includegraphics[height=2.1cm]{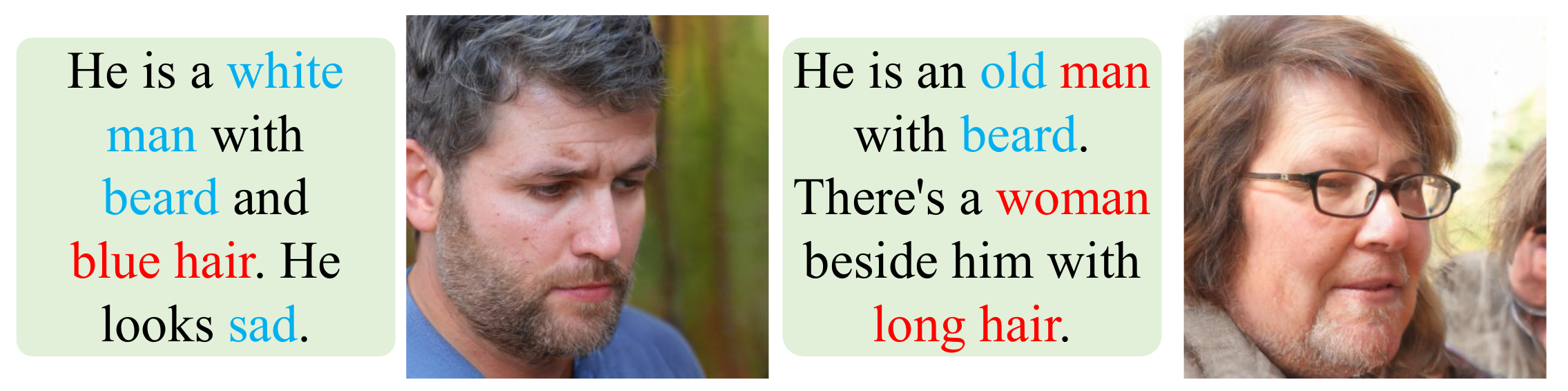}
    \caption{The failure cases. The successful attributes are \textcolor[RGB]{0, 176, 240}{blue} and wrong attributes are \textcolor[RGB]{255, 0, 0}{red} (best view in color).}
    \label{fig:Failure}
\end{figure}

\subsection{Failure Cases and Discussions}
\label{subsec:Failure}

There are two kinds of failure cases of the proposed method. First, if the image we expect is out of the generation domain of StyleGAN, it is hard to be generated. Second, if there is more than one person described in one input, our framework may be confused. These failure cases are shown in Figure \ref{fig:Failure}.

We consider that there may be two reasons for these failures. First, there's a limitation of the generation domain of StyleGAN. Second, there's still room for improvement of our framework to better leverage CLIP embeddings of texts with complicated texts.

\section{Conclusions}
\label{sec:Conclusions}

We have proposed a framework for illustrating complicated semantics. Our framework is able to deal with various text inputs and generate impressive images with high quality, fidelity, and semantic alignment to the input texts due to our appropriate design with pre-trained models including CLIP and our StyleGAN. Our method is a general method that can be used to translate multiple kinds of objectives, \eg, human faces, churches, and cats. Extensive experiments on different kinds of input text descriptions demonstrate the superiority of our method. User study also shows that most people prefer our method to other state-of-the-art methods because of the visually pleasant results and semantic accurateness. 

\clearpage
\bibliographystyle{ACM-Reference-Format}
\balance
\bibliography{release}


\begin{thebibliography}{48}


\ifx \showCODEN    \undefined \def \showCODEN     #1{\unskip}     \fi
\ifx \showDOI      \undefined \def \showDOI       #1{#1}\fi
\ifx \showISBNx    \undefined \def \showISBNx     #1{\unskip}     \fi
\ifx \showISBNxiii \undefined \def \showISBNxiii  #1{\unskip}     \fi
\ifx \showISSN     \undefined \def \showISSN      #1{\unskip}     \fi
\ifx \showLCCN     \undefined \def \showLCCN      #1{\unskip}     \fi
\ifx \shownote     \undefined \def \shownote      #1{#1}          \fi
\ifx \showarticletitle \undefined \def \showarticletitle #1{#1}   \fi
\ifx \showURL      \undefined \def \showURL       {\relax}        \fi
\providecommand\bibfield[2]{#2}
\providecommand\bibinfo[2]{#2}
\providecommand\natexlab[1]{#1}
\providecommand\showeprint[2][]{arXiv:#2}

\bibitem[Devlin et~al\mbox{.}(2018)]%
        {BERT}
\bibfield{author}{\bibinfo{person}{Jacob Devlin}, \bibinfo{person}{Ming-Wei
  Chang}, \bibinfo{person}{Kenton Lee}, {and} \bibinfo{person}{Kristina
  Toutanova}.} \bibinfo{year}{2018}\natexlab{}.
\newblock \showarticletitle{BERT: Pre-training of Deep Bidirectional
  Transformers for Language Understanding}.
\newblock \bibinfo{journal}{\emph{arXiv preprint arXiv:1810.04805}}
  (\bibinfo{year}{2018}).
\newblock


\bibitem[Dong et~al\mbox{.}(2017)]%
        {SISGAN}
\bibfield{author}{\bibinfo{person}{Hao Dong}, \bibinfo{person}{Simiao Yu},
  \bibinfo{person}{Chao Wu}, {and} \bibinfo{person}{Yike Guo}.}
  \bibinfo{year}{2017}\natexlab{}.
\newblock \showarticletitle{Semantic Image Synthesis via Adversarial Learning}.
  In \bibinfo{booktitle}{\emph{Proceedings of the IEEE International Conference
  on Computer Vision}}. \bibinfo{pages}{5706--5714}.
\newblock


\bibitem[Dosovitskiy et~al\mbox{.}(2020)]%
        {ViT}
\bibfield{author}{\bibinfo{person}{Alexey Dosovitskiy}, \bibinfo{person}{Lucas
  Beyer}, \bibinfo{person}{Alexander Kolesnikov}, \bibinfo{person}{Dirk
  Weissenborn}, \bibinfo{person}{Xiaohua Zhai}, \bibinfo{person}{Thomas
  Unterthiner}, \bibinfo{person}{Mostafa Dehghani}, \bibinfo{person}{Matthias
  Minderer}, \bibinfo{person}{Georg Heigold}, \bibinfo{person}{Sylvain Gelly},
  {et~al\mbox{.}}} \bibinfo{year}{2020}\natexlab{}.
\newblock \showarticletitle{An Image is Worth 16x16 Words: Transformers for
  Image Recognition at Scale}.
\newblock \bibinfo{journal}{\emph{arXiv preprint arXiv:2010.11929}}
  (\bibinfo{year}{2020}).
\newblock


\bibitem[Fu et~al\mbox{.}(2017)]%
        {LookCloser}
\bibfield{author}{\bibinfo{person}{Jianlong Fu}, \bibinfo{person}{Heliang
  Zheng}, {and} \bibinfo{person}{Tao Mei}.} \bibinfo{year}{2017}\natexlab{}.
\newblock \showarticletitle{Look Closer to See Better: Recurrent Attention
  Convolutional Neural Network for Fine-grained Image Recognition}. In
  \bibinfo{booktitle}{\emph{Proceedings of the IEEE Conference on Computer
  Vision and Pattern Recognition}}. \bibinfo{pages}{4438--4446}.
\newblock


\bibitem[Gal et~al\mbox{.}(2021)]%
        {StyleGAN-nada}
\bibfield{author}{\bibinfo{person}{Rinon Gal}, \bibinfo{person}{Or Patashnik},
  \bibinfo{person}{Haggai Maron}, \bibinfo{person}{Gal Chechik}, {and}
  \bibinfo{person}{Daniel Cohen-Or}.} \bibinfo{year}{2021}\natexlab{}.
\newblock \showarticletitle{StyleGAN-NADA: CLIP-Guided Domain Adaptation of
  Image Generators}.
\newblock \bibinfo{journal}{\emph{arXiv preprint arXiv:2108.00946}}
  (\bibinfo{year}{2021}).
\newblock


\bibitem[Goodfellow et~al\mbox{.}(2014)]%
        {GAN}
\bibfield{author}{\bibinfo{person}{Ian Goodfellow}, \bibinfo{person}{Jean
  Pouget-Abadie}, \bibinfo{person}{Mehdi Mirza}, \bibinfo{person}{Bing Xu},
  \bibinfo{person}{David Warde-Farley}, \bibinfo{person}{Sherjil Ozair},
  \bibinfo{person}{Aaron Courville}, {and} \bibinfo{person}{Yoshua Bengio}.}
  \bibinfo{year}{2014}\natexlab{}.
\newblock \showarticletitle{Generative Adversarial Nets}.
\newblock \bibinfo{journal}{\emph{Advances in Neural Information Processing
  Systems}}  \bibinfo{volume}{27} (\bibinfo{year}{2014}).
\newblock


\bibitem[He et~al\mbox{.}(2015)]%
        {PRelu}
\bibfield{author}{\bibinfo{person}{Kaiming He}, \bibinfo{person}{Xiangyu
  Zhang}, \bibinfo{person}{Shaoqing Ren}, {and} \bibinfo{person}{Jian Sun}.}
  \bibinfo{year}{2015}\natexlab{}.
\newblock \showarticletitle{Delving Deep into Rectifiers: Surpassing
  Human-Level Performance on Imagenet Classification}. In
  \bibinfo{booktitle}{\emph{Proceedings of the IEEE International Conference on
  Computer Vision}}. \bibinfo{pages}{1026--1034}.
\newblock


\bibitem[Heusel et~al\mbox{.}(2017)]%
        {TTUR}
\bibfield{author}{\bibinfo{person}{Martin Heusel}, \bibinfo{person}{Hubert
  Ramsauer}, \bibinfo{person}{Thomas Unterthiner}, \bibinfo{person}{Bernhard
  Nessler}, {and} \bibinfo{person}{Sepp Hochreiter}.}
  \bibinfo{year}{2017}\natexlab{}.
\newblock \showarticletitle{GANs Trained by a Two Time-Scale Update Rule
  Converge to a Local Nash Equilibrium}.
\newblock \bibinfo{journal}{\emph{Advances in Neural Information Processing
  Systems}}  \bibinfo{volume}{30} (\bibinfo{year}{2017}).
\newblock


\bibitem[Ioffe and Szegedy(2015)]%
        {BatchNorm}
\bibfield{author}{\bibinfo{person}{Sergey Ioffe} {and}
  \bibinfo{person}{Christian Szegedy}.} \bibinfo{year}{2015}\natexlab{}.
\newblock \showarticletitle{Batch Normalization: Accelerating Deep Network
  Training by Reducing Internal Covariate Shift}. In
  \bibinfo{booktitle}{\emph{Proceedings of the International Conference on
  Machine Learning}}. \bibinfo{pages}{448--456}.
\newblock


\bibitem[Karras et~al\mbox{.}(2019)]%
        {StyleGANv1}
\bibfield{author}{\bibinfo{person}{Tero Karras}, \bibinfo{person}{Samuli
  Laine}, {and} \bibinfo{person}{Timo Aila}.} \bibinfo{year}{2019}\natexlab{}.
\newblock \showarticletitle{A Style-Based Generator Architecture for Generative
  Adversarial Networks}. In \bibinfo{booktitle}{\emph{Proceedings of the
  IEEE/CVF Conference on Computer Vision and Pattern Recognition}}.
  \bibinfo{pages}{4401--4410}.
\newblock


\bibitem[Karras et~al\mbox{.}(2020)]%
        {StyleGANv2}
\bibfield{author}{\bibinfo{person}{Tero Karras}, \bibinfo{person}{Samuli
  Laine}, \bibinfo{person}{Miika Aittala}, \bibinfo{person}{Janne Hellsten},
  \bibinfo{person}{Jaakko Lehtinen}, {and} \bibinfo{person}{Timo Aila}.}
  \bibinfo{year}{2020}\natexlab{}.
\newblock \showarticletitle{Analyzing and Improving the Image Quality of
  StyleGAN}. In \bibinfo{booktitle}{\emph{Proceedings of the IEEE/CVF
  Conference on Computer Vision and Pattern Recognition}}.
  \bibinfo{pages}{8110--8119}.
\newblock


\bibitem[Kingma and Ba(2014)]%
        {Adam}
\bibfield{author}{\bibinfo{person}{Diederik~P Kingma} {and}
  \bibinfo{person}{Jimmy Ba}.} \bibinfo{year}{2014}\natexlab{}.
\newblock \showarticletitle{Adam: A Method for Stochastic Optimization}.
\newblock \bibinfo{journal}{\emph{arXiv preprint arXiv:1412.6980}}
  (\bibinfo{year}{2014}).
\newblock


\bibitem[Li et~al\mbox{.}(2019a)]%
        {ControlGAN}
\bibfield{author}{\bibinfo{person}{Bowen Li}, \bibinfo{person}{Xiaojuan Qi},
  \bibinfo{person}{Thomas Lukasiewicz}, {and} \bibinfo{person}{Philip Torr}.}
  \bibinfo{year}{2019}\natexlab{a}.
\newblock \showarticletitle{Controllable Text-to-Image Generation}.
\newblock \bibinfo{journal}{\emph{Advances in Neural Information Processing
  Systems}}  \bibinfo{volume}{32} (\bibinfo{year}{2019}).
\newblock


\bibitem[Li et~al\mbox{.}(2020a)]%
        {ManiGAN}
\bibfield{author}{\bibinfo{person}{Bowen Li}, \bibinfo{person}{Xiaojuan Qi},
  \bibinfo{person}{Thomas Lukasiewicz}, {and} \bibinfo{person}{Philip~HS
  Torr}.} \bibinfo{year}{2020}\natexlab{a}.
\newblock \showarticletitle{ManiGAN: Text-Guided Image Manipulation}. In
  \bibinfo{booktitle}{\emph{Proceedings of the IEEE/CVF Conference on Computer
  Vision and Pattern Recognition}}. \bibinfo{pages}{7880--7889}.
\newblock


\bibitem[Li et~al\mbox{.}(2020b)]%
        {Lightweight-Manipulation}
\bibfield{author}{\bibinfo{person}{Bowen Li}, \bibinfo{person}{Xiaojuan Qi},
  \bibinfo{person}{Philip Torr}, {and} \bibinfo{person}{Thomas Lukasiewicz}.}
  \bibinfo{year}{2020}\natexlab{b}.
\newblock \showarticletitle{Lightweight Generative Adversarial Networks for
  Text-Guided Image Manipulation}.
\newblock \bibinfo{journal}{\emph{Advances in Neural Information Processing
  Systems}}  \bibinfo{volume}{33} (\bibinfo{year}{2020}),
  \bibinfo{pages}{22020--22031}.
\newblock


\bibitem[Li et~al\mbox{.}(2019b)]%
        {Obj-GAN}
\bibfield{author}{\bibinfo{person}{Wenbo Li}, \bibinfo{person}{Pengchuan
  Zhang}, \bibinfo{person}{Lei Zhang}, \bibinfo{person}{Qiuyuan Huang},
  \bibinfo{person}{Xiaodong He}, \bibinfo{person}{Siwei Lyu}, {and}
  \bibinfo{person}{Jianfeng Gao}.} \bibinfo{year}{2019}\natexlab{b}.
\newblock \showarticletitle{Object-driven Text-to-Image Synthesis via
  Adversarial Training}. In \bibinfo{booktitle}{\emph{Proceedings of the
  IEEE/CVF Conference on Computer Vision and Pattern Recognition}}.
  \bibinfo{pages}{12174--12182}.
\newblock


\bibitem[Li and Liang(2021)]%
        {OptimizePrompt}
\bibfield{author}{\bibinfo{person}{Xiang~Lisa Li} {and} \bibinfo{person}{Percy
  Liang}.} \bibinfo{year}{2021}\natexlab{}.
\newblock \showarticletitle{Prefix-Tuning: Optimizing Continuous Prompts for
  Generation}.
\newblock \bibinfo{journal}{\emph{arXiv preprint arXiv:2101.00190}}
  (\bibinfo{year}{2021}).
\newblock


\bibitem[Liu et~al\mbox{.}(2021a)]%
        {prompt-survey}
\bibfield{author}{\bibinfo{person}{Pengfei Liu}, \bibinfo{person}{Weizhe Yuan},
  \bibinfo{person}{Jinlan Fu}, \bibinfo{person}{Zhengbao Jiang},
  \bibinfo{person}{Hiroaki Hayashi}, {and} \bibinfo{person}{Graham Neubig}.}
  \bibinfo{year}{2021}\natexlab{a}.
\newblock \showarticletitle{Pre-train, Prompt, and Predict: A Systematic Survey
  of Prompting Methods in Natural Language Processing}.
\newblock \bibinfo{journal}{\emph{arXiv preprint arXiv:2107.13586}}
  (\bibinfo{year}{2021}).
\newblock


\bibitem[Liu et~al\mbox{.}(2021b)]%
        {GPTUnderstands}
\bibfield{author}{\bibinfo{person}{Xiao Liu}, \bibinfo{person}{Yanan Zheng},
  \bibinfo{person}{Zhengxiao Du}, \bibinfo{person}{Ming Ding},
  \bibinfo{person}{Yujie Qian}, \bibinfo{person}{Zhilin Yang}, {and}
  \bibinfo{person}{Jie Tang}.} \bibinfo{year}{2021}\natexlab{b}.
\newblock \showarticletitle{GPT Understands, Too}.
\newblock \bibinfo{journal}{\emph{arXiv preprint arXiv:2103.10385}}
  (\bibinfo{year}{2021}).
\newblock


\bibitem[Mirza and Osindero(2014)]%
        {cGAN}
\bibfield{author}{\bibinfo{person}{Mehdi Mirza} {and} \bibinfo{person}{Simon
  Osindero}.} \bibinfo{year}{2014}\natexlab{}.
\newblock \showarticletitle{Conditional Generative Adversarial Nets}.
\newblock \bibinfo{journal}{\emph{arXiv preprint arXiv:1411.1784}}
  (\bibinfo{year}{2014}).
\newblock


\bibitem[Patashnik et~al\mbox{.}(2021)]%
        {StyleCLIP}
\bibfield{author}{\bibinfo{person}{Or Patashnik}, \bibinfo{person}{Zongze Wu},
  \bibinfo{person}{Eli Shechtman}, \bibinfo{person}{Daniel Cohen-Or}, {and}
  \bibinfo{person}{Dani Lischinski}.} \bibinfo{year}{2021}\natexlab{}.
\newblock \showarticletitle{StyleCLIP: Text-Driven Manipulation of StyleGAN
  Imagery}. In \bibinfo{booktitle}{\emph{Proceedings of the IEEE/CVF
  International Conference on Computer Vision}}. \bibinfo{pages}{2085--2094}.
\newblock


\bibitem[Qiao et~al\mbox{.}(2019)]%
        {MirrorGAN}
\bibfield{author}{\bibinfo{person}{Tingting Qiao}, \bibinfo{person}{Jing
  Zhang}, \bibinfo{person}{Duanqing Xu}, {and} \bibinfo{person}{Dacheng Tao}.}
  \bibinfo{year}{2019}\natexlab{}.
\newblock \showarticletitle{MirrorGAN: Learning Text-to-Image Generation by
  Redescription}. In \bibinfo{booktitle}{\emph{Proceedings of the IEEE/CVF
  Conference on Computer Vision and Pattern Recognition}}.
  \bibinfo{pages}{1505--1514}.
\newblock


\bibitem[Radford et~al\mbox{.}(2021)]%
        {CLIP}
\bibfield{author}{\bibinfo{person}{Alec Radford}, \bibinfo{person}{Jong~Wook
  Kim}, \bibinfo{person}{Chris Hallacy}, \bibinfo{person}{Aditya Ramesh},
  \bibinfo{person}{Gabriel Goh}, \bibinfo{person}{Sandhini Agarwal},
  \bibinfo{person}{Girish Sastry}, \bibinfo{person}{Amanda Askell},
  \bibinfo{person}{Pamela Mishkin}, \bibinfo{person}{Jack Clark},
  {et~al\mbox{.}}} \bibinfo{year}{2021}\natexlab{}.
\newblock \showarticletitle{Learning Transferable Visual Models from Natural
  Language Supervision}. In \bibinfo{booktitle}{\emph{International Conference
  on Machine Learning}}. \bibinfo{pages}{8748--8763}.
\newblock


\bibitem[Ramesh et~al\mbox{.}(2021)]%
        {DALL-E}
\bibfield{author}{\bibinfo{person}{Aditya Ramesh}, \bibinfo{person}{Mikhail
  Pavlov}, \bibinfo{person}{Gabriel Goh}, \bibinfo{person}{Scott Gray},
  \bibinfo{person}{Chelsea Voss}, \bibinfo{person}{Alec Radford},
  \bibinfo{person}{Mark Chen}, {and} \bibinfo{person}{Ilya Sutskever}.}
  \bibinfo{year}{2021}\natexlab{}.
\newblock \showarticletitle{Zero-Shot Text-to-Image Generation}. In
  \bibinfo{booktitle}{\emph{International Conference on Machine Learning}}.
  \bibinfo{pages}{8821--8831}.
\newblock


\bibitem[Reed et~al\mbox{.}(2016)]%
        {GAN-INT-CLS}
\bibfield{author}{\bibinfo{person}{Scott Reed}, \bibinfo{person}{Zeynep Akata},
  \bibinfo{person}{Xinchen Yan}, \bibinfo{person}{Lajanugen Logeswaran},
  \bibinfo{person}{Bernt Schiele}, {and} \bibinfo{person}{Honglak Lee}.}
  \bibinfo{year}{2016}\natexlab{}.
\newblock \showarticletitle{Generative Adversarial Text to Image Synthesis}. In
  \bibinfo{booktitle}{\emph{International Conference on Machine Learning}}.
  \bibinfo{pages}{1060--1069}.
\newblock


\bibitem[Reynolds and McDonell(2021)]%
        {PromptProgramming}
\bibfield{author}{\bibinfo{person}{Laria Reynolds} {and} \bibinfo{person}{Kyle
  McDonell}.} \bibinfo{year}{2021}\natexlab{}.
\newblock \showarticletitle{Prompt Programming for Large Language Models:
  Beyond the Few-Shot Paradigm}. In \bibinfo{booktitle}{\emph{Extended
  Abstracts of the 2021 CHI Conference on Human Factors in Computing Systems}}.
  \bibinfo{pages}{1--7}.
\newblock


\bibitem[Salimans et~al\mbox{.}(2016)]%
        {InceptionScore}
\bibfield{author}{\bibinfo{person}{Tim Salimans}, \bibinfo{person}{Ian
  Goodfellow}, \bibinfo{person}{Wojciech Zaremba}, \bibinfo{person}{Vicki
  Cheung}, \bibinfo{person}{Alec Radford}, {and} \bibinfo{person}{Xi Chen}.}
  \bibinfo{year}{2016}\natexlab{}.
\newblock \showarticletitle{Improved Techniques for Training GANs}.
\newblock \bibinfo{journal}{\emph{Advances in Neural Information Processing
  Systems}}  \bibinfo{volume}{29} (\bibinfo{year}{2016}).
\newblock


\bibitem[Schick and Sch{\"u}tze(2020)]%
        {PromptBasic1}
\bibfield{author}{\bibinfo{person}{Timo Schick} {and} \bibinfo{person}{Hinrich
  Sch{\"u}tze}.} \bibinfo{year}{2020}\natexlab{}.
\newblock \showarticletitle{It's Not Just Size That Matters: Small Language
  Models Are Also Few-Shot Learners}.
\newblock \bibinfo{journal}{\emph{arXiv preprint arXiv:2009.07118}}
  (\bibinfo{year}{2020}).
\newblock


\bibitem[Shin et~al\mbox{.}(2020)]%
        {Autoprompt}
\bibfield{author}{\bibinfo{person}{Taylor Shin}, \bibinfo{person}{Yasaman
  Razeghi}, \bibinfo{person}{Robert~L Logan~IV}, \bibinfo{person}{Eric
  Wallace}, {and} \bibinfo{person}{Sameer Singh}.}
  \bibinfo{year}{2020}\natexlab{}.
\newblock \showarticletitle{AutoPrompt: Eliciting Knowledge from Language
  Models with Automatically Generated Prompts}.
\newblock \bibinfo{journal}{\emph{arXiv preprint arXiv:2010.15980}}
  (\bibinfo{year}{2020}).
\newblock


\bibitem[Srivastava et~al\mbox{.}(2014)]%
        {Dropout}
\bibfield{author}{\bibinfo{person}{Nitish Srivastava},
  \bibinfo{person}{Geoffrey Hinton}, \bibinfo{person}{Alex Krizhevsky},
  \bibinfo{person}{Ilya Sutskever}, {and} \bibinfo{person}{Ruslan
  Salakhutdinov}.} \bibinfo{year}{2014}\natexlab{}.
\newblock \showarticletitle{Dropout: A Simple Way to Prevent Neural Networks
  from Overfitting}.
\newblock \bibinfo{journal}{\emph{The Journal of Machine Learning Research}}
  \bibinfo{volume}{15}, \bibinfo{number}{1} (\bibinfo{year}{2014}),
  \bibinfo{pages}{1929--1958}.
\newblock


\bibitem[Tao et~al\mbox{.}(2020)]%
        {DF-GAN}
\bibfield{author}{\bibinfo{person}{Ming Tao}, \bibinfo{person}{Hao Tang},
  \bibinfo{person}{Songsong Wu}, \bibinfo{person}{Nicu Sebe},
  \bibinfo{person}{Xiao-Yuan Jing}, \bibinfo{person}{Fei Wu}, {and}
  \bibinfo{person}{Bingkun Bao}.} \bibinfo{year}{2020}\natexlab{}.
\newblock \showarticletitle{DF-GAN: Deep Fusion Generative Adversarial Networks
  for Text-to-Image Synthesis}.
\newblock \bibinfo{journal}{\emph{arXiv preprint arXiv:2008.05865}}
  (\bibinfo{year}{2020}).
\newblock


\bibitem[Vaswani et~al\mbox{.}(2017)]%
        {Attention}
\bibfield{author}{\bibinfo{person}{Ashish Vaswani}, \bibinfo{person}{Noam
  Shazeer}, \bibinfo{person}{Niki Parmar}, \bibinfo{person}{Jakob Uszkoreit},
  \bibinfo{person}{Llion Jones}, \bibinfo{person}{Aidan~N Gomez},
  \bibinfo{person}{{\L}ukasz Kaiser}, {and} \bibinfo{person}{Illia
  Polosukhin}.} \bibinfo{year}{2017}\natexlab{}.
\newblock \showarticletitle{Attention is All You Need}.
\newblock \bibinfo{journal}{\emph{Advances in Neural Information Processing
  Systems}}  \bibinfo{volume}{30} (\bibinfo{year}{2017}).
\newblock


\bibitem[Wang and Yu(2020)]%
        {Cartoonize}
\bibfield{author}{\bibinfo{person}{Xinrui Wang} {and} \bibinfo{person}{Jinze
  Yu}.} \bibinfo{year}{2020}\natexlab{}.
\newblock \showarticletitle{Learning to Cartoonize Using White-box Cartoon
  Representations}. In \bibinfo{booktitle}{\emph{Proceedings of the IEEE/CVF
  Conference on Computer Vision and Pattern Recognition}}.
  \bibinfo{pages}{8090--8099}.
\newblock


\bibitem[Xia et~al\mbox{.}(2021a)]%
        {TediGAN-A}
\bibfield{author}{\bibinfo{person}{Weihao Xia}, \bibinfo{person}{Yujiu Yang},
  \bibinfo{person}{Jing-Hao Xue}, {and} \bibinfo{person}{Baoyuan Wu}.}
  \bibinfo{year}{2021}\natexlab{a}.
\newblock \showarticletitle{TediGAN: Text-Guided Diverse Face Image Generation
  and Manipulation}. In \bibinfo{booktitle}{\emph{Proceedings of the IEEE/CVF
  Conference on Computer Vision and Pattern Recognition}}.
  \bibinfo{pages}{2256--2265}.
\newblock


\bibitem[Xia et~al\mbox{.}(2021b)]%
        {TediGAN-B}
\bibfield{author}{\bibinfo{person}{Weihao Xia}, \bibinfo{person}{Yujiu Yang},
  \bibinfo{person}{Jing-Hao Xue}, {and} \bibinfo{person}{Baoyuan Wu}.}
  \bibinfo{year}{2021}\natexlab{b}.
\newblock \showarticletitle{Towards Open-World Text-Guided Face Image
  Generation and Manipulation}.
\newblock \bibinfo{journal}{\emph{arXiv preprint arXiv: 2104.08910}}
  (\bibinfo{year}{2021}).
\newblock


\bibitem[Xu et~al\mbox{.}(2018)]%
        {AttnGAN}
\bibfield{author}{\bibinfo{person}{Tao Xu}, \bibinfo{person}{Pengchuan Zhang},
  \bibinfo{person}{Qiuyuan Huang}, \bibinfo{person}{Han Zhang},
  \bibinfo{person}{Zhe Gan}, \bibinfo{person}{Xiaolei Huang}, {and}
  \bibinfo{person}{Xiaodong He}.} \bibinfo{year}{2018}\natexlab{}.
\newblock \showarticletitle{Attngan: Fine-grained text to image generation with
  attentional generative adversarial networks}. In
  \bibinfo{booktitle}{\emph{Proceedings of the IEEE Conference on Computer
  Vision and Pattern Recognition}}. \bibinfo{pages}{1316--1324}.
\newblock


\bibitem[Yang et~al\mbox{.}(2020)]%
        {TTSR}
\bibfield{author}{\bibinfo{person}{Fuzhi Yang}, \bibinfo{person}{Huan Yang},
  \bibinfo{person}{Jianlong Fu}, \bibinfo{person}{Hongtao Lu}, {and}
  \bibinfo{person}{Baining Guo}.} \bibinfo{year}{2020}\natexlab{}.
\newblock \showarticletitle{Learning Texture Transformer Network for Image
  Super-Resolution}. In \bibinfo{booktitle}{\emph{Proceedings of the IEEE/CVF
  Conference on Computer Vision and Pattern Recognition}}.
  \bibinfo{pages}{5791--5800}.
\newblock


\bibitem[Yin et~al\mbox{.}(2019)]%
        {SD-GAN}
\bibfield{author}{\bibinfo{person}{Guojun Yin}, \bibinfo{person}{Bin Liu},
  \bibinfo{person}{Lu Sheng}, \bibinfo{person}{Nenghai Yu},
  \bibinfo{person}{Xiaogang Wang}, {and} \bibinfo{person}{Jing Shao}.}
  \bibinfo{year}{2019}\natexlab{}.
\newblock \showarticletitle{Semantics Disentangling for Text-to-Image
  Generation}. In \bibinfo{booktitle}{\emph{Proceedings of the IEEE/CVF
  Conference on Computer Vision and Pattern Recognition}}.
  \bibinfo{pages}{2327--2336}.
\newblock


\bibitem[Zeng et~al\mbox{.}(2020)]%
        {STTN}
\bibfield{author}{\bibinfo{person}{Yanhong Zeng}, \bibinfo{person}{Jianlong
  Fu}, {and} \bibinfo{person}{Hongyang Chao}.} \bibinfo{year}{2020}\natexlab{}.
\newblock \showarticletitle{Learning Joint Spatial-Temporal Transformations for
  Video Inpainting}. In \bibinfo{booktitle}{\emph{Proceedings of the European
  Conference on Computer Vision}}. \bibinfo{pages}{528--543}.
\newblock


\bibitem[Zeng et~al\mbox{.}(2019)]%
        {PEN-Net}
\bibfield{author}{\bibinfo{person}{Yanhong Zeng}, \bibinfo{person}{Jianlong
  Fu}, \bibinfo{person}{Hongyang Chao}, {and} \bibinfo{person}{Baining Guo}.}
  \bibinfo{year}{2019}\natexlab{}.
\newblock \showarticletitle{Learning Pyramid-Context Encoder Network for
  High-Quality Image Inpainting}. In \bibinfo{booktitle}{\emph{Proceedings of
  the IEEE/CVF Conference on Computer Vision and Pattern Recognition}}.
  \bibinfo{pages}{1486--1494}.
\newblock


\bibitem[Zeng et~al\mbox{.}(2022)]%
        {AOT-GAN}
\bibfield{author}{\bibinfo{person}{Yanhong Zeng}, \bibinfo{person}{Jianlong
  Fu}, \bibinfo{person}{Hongyang Chao}, {and} \bibinfo{person}{Baining Guo}.}
  \bibinfo{year}{2022}\natexlab{}.
\newblock \showarticletitle{Aggregated Contextual Transformations for
  High-Resolution Image Inpainting}.
\newblock \bibinfo{journal}{\emph{IEEE Transactions on Visualization and
  Computer Graphics}} (\bibinfo{year}{2022}).
\newblock


\bibitem[Zeng et~al\mbox{.}(2021)]%
        {ImproveVisualQuality}
\bibfield{author}{\bibinfo{person}{Yanhong Zeng}, \bibinfo{person}{Huan Yang},
  \bibinfo{person}{Hongyang Chao}, \bibinfo{person}{Jianbo Wang}, {and}
  \bibinfo{person}{Jianlong Fu}.} \bibinfo{year}{2021}\natexlab{}.
\newblock \showarticletitle{Improving Visual Quality of Image Synthesis by a
  Token-based Generator with Transformers}.
\newblock \bibinfo{journal}{\emph{Advances in Neural Information Processing
  Systems}}  \bibinfo{volume}{34} (\bibinfo{year}{2021}),
  \bibinfo{pages}{21125--21137}.
\newblock


\bibitem[Zhang et~al\mbox{.}(2017)]%
        {StackGAN}
\bibfield{author}{\bibinfo{person}{Han Zhang}, \bibinfo{person}{Tao Xu},
  \bibinfo{person}{Hongsheng Li}, \bibinfo{person}{Shaoting Zhang},
  \bibinfo{person}{Xiaogang Wang}, \bibinfo{person}{Xiaolei Huang}, {and}
  \bibinfo{person}{Dimitris~N Metaxas}.} \bibinfo{year}{2017}\natexlab{}.
\newblock \showarticletitle{StackGAN: Text to Photo-realistic Image Synthesis
  with Stacked Generative Adversarial Networks}. In
  \bibinfo{booktitle}{\emph{Proceedings of the IEEE International Conference on
  Computer Vision}}. \bibinfo{pages}{5907--5915}.
\newblock


\bibitem[Zhang et~al\mbox{.}(2018b)]%
        {StackGAN++}
\bibfield{author}{\bibinfo{person}{Han Zhang}, \bibinfo{person}{Tao Xu},
  \bibinfo{person}{Hongsheng Li}, \bibinfo{person}{Shaoting Zhang},
  \bibinfo{person}{Xiaogang Wang}, \bibinfo{person}{Xiaolei Huang}, {and}
  \bibinfo{person}{Dimitris~N Metaxas}.} \bibinfo{year}{2018}\natexlab{b}.
\newblock \showarticletitle{StackGAN++: Realistic Image Synthesis with Stacked
  Generative Adversarial Networks}.
\newblock \bibinfo{journal}{\emph{IEEE Transactions on Pattern Analysis and
  Machine Intelligence}} \bibinfo{volume}{41}, \bibinfo{number}{8}
  (\bibinfo{year}{2018}), \bibinfo{pages}{1947--1962}.
\newblock


\bibitem[Zhang et~al\mbox{.}(2022)]%
        {MetroGAN}
\bibfield{author}{\bibinfo{person}{Weiyu Zhang}, \bibinfo{person}{Yiyang Ma},
  \bibinfo{person}{Di Zhu}, \bibinfo{person}{Lei Dong}, {and}
  \bibinfo{person}{Yu Liu}.} \bibinfo{year}{2022}\natexlab{}.
\newblock \showarticletitle{MetroGAN: Simulating Urban Morphology with
  Generative Adversarial Network}.
\newblock \bibinfo{journal}{\emph{arXiv preprint arXiv:2207.02590}}
  (\bibinfo{year}{2022}).
\newblock


\bibitem[Zhang et~al\mbox{.}(2018a)]%
        {HD-GAN}
\bibfield{author}{\bibinfo{person}{Zizhao Zhang}, \bibinfo{person}{Yuanpu Xie},
  {and} \bibinfo{person}{Lin Yang}.} \bibinfo{year}{2018}\natexlab{a}.
\newblock \showarticletitle{Photographic Text-to-Image Synthesis with a
  Hierarchically-nested Adversarial Network}. In
  \bibinfo{booktitle}{\emph{Proceedings of the IEEE Conference on Computer
  Vision and Pattern Recognition}}. \bibinfo{pages}{6199--6208}.
\newblock


\bibitem[Zheng et~al\mbox{.}(2017)]%
        {LearningMultiAttention}
\bibfield{author}{\bibinfo{person}{Heliang Zheng}, \bibinfo{person}{Jianlong
  Fu}, \bibinfo{person}{Tao Mei}, {and} \bibinfo{person}{Jiebo Luo}.}
  \bibinfo{year}{2017}\natexlab{}.
\newblock \showarticletitle{Learning Multi-Attention Convolutional Neural
  Network for Fine-grained Image Recognition}. In
  \bibinfo{booktitle}{\emph{Proceedings of the IEEE International Conference on
  Computer Vision}}. \bibinfo{pages}{5209--5217}.
\newblock


\bibitem[Zhu et~al\mbox{.}(2019)]%
        {DM-GAN}
\bibfield{author}{\bibinfo{person}{Minfeng Zhu}, \bibinfo{person}{Pingbo Pan},
  \bibinfo{person}{Wei Chen}, {and} \bibinfo{person}{Yi Yang}.}
  \bibinfo{year}{2019}\natexlab{}.
\newblock \showarticletitle{DM-GAN: Dynamic Memory Generative Adversarial
  Networks for Text-to-Image Synthesis}. In
  \bibinfo{booktitle}{\emph{Proceedings of the IEEE/CVF Conference on Computer
  Vision and Pattern Recognition}}. \bibinfo{pages}{5802--5810}.
\newblock


\end{thebibliography}

\clearpage

\appendix

\section{Implementation Details}
\label{sec:Implementation}
In this section, we introduce all the implementation details in order to ensure the reproducibility of our results.
We use CLIP \cite{CLIP} with ViT-B/32 \cite{ViT} to semantically align images and texts in our work. It should be noted that the lengths of all the $CTE$s and $CIE$s in our work are normalized because only the orientations of these embeddings contain semantics. In practice, there is a trick that we normalize their lengths to $\sqrt{512}$ instead of 1. That is because we project $CIE_{input}$ to $SE$ in the second module in our work and the length expectation of $SE$ is $\sqrt{512}$. $SE$s are embeddings of StyleGAN \cite{StyleGANv1, StyleGANv2} $\mathcal{Z}$ space with 512 dimensions, and the distribution of this space is a standard normal distribution. So, if the length of $CIE_{input}$ is close to the length of the expected $SE$ output, it will be easy for the network to converge. 

Since there are two major modules in our framework, we introduce the details of them.
The first module of our method is only a combination of linear operations, so there are not any additional details to introduce. The specific method of getting \textit{prompt embedding}s we use in this section is shown below. In order to get $CIE_{prompt}$, we sample 150,000 $SE$s in StyleGAN $\mathcal{Z}$ space, generate images and extract $CIE$s by CLIP from them. As we discuss in Section 3.1 of the main paper, the $CIE_{prompt}$ is the average value of these $CIE$s. These $CIE$-$SE$ pairs are also used to train the network in the second module. In the aspect of $CTE_{prompt}$, we extract $CTE$s from some certain sentences as $CTE_{prompt}$ for different tasks. These sentences are shown in Table \ref{tab:TextPrompt}.

The second module uses a fully connected network with dense connections called $C2S$ projection network to project $CIE_{input}$ which is got in the first module to the corresponding $SE$. This network needs $CIE$-$SE$ pairs to train and these pairs are the data which is used to get $CIE_{prompt}$s above. The $C2S$ projection network has 5 dense blocks as body part, 2 fully connected layers as head part, and tail part. Dropout layers \cite{Dropout} with a dropout ratio of 0.1 are applied after every dense block and skip connections are applied in parallel with each dense block. For every single dense block, there are 10 fully connected layers and PReLU \cite{PRelu} are adopted as activation functions after every fully connected layer. BatchNorm \cite{BatchNorm} layers are also applied in dense blocks. The architecture of $C2S$ projection network and dense block is shown in Table \ref{tab:C2S_Arch} and Table \ref{tab:DenseBlock} respectively. We train the $C2S$ projection network for 150,000 iterations with batch size of 16 on two M40 GPUs. We use Adam \cite{Adam} as our optimizer and the initial learning rate is set to $1\times10^{-4}$. We also adopt cosine annealing to the learning rate and it will drop to $1\times10^{-7}$ monotonically at the last iteration. After getting the $SE$, we generate realistic image from it by StyleGAN2 \cite{StyleGANv2}.

\begin{table}[htbp]
    \centering
    \begin{tabular}{p{3cm}p{3cm}}
        \toprule
        Translation Task & Text Prompt Sentence
        \\
        \midrule
        Text-to-Human Face & A normal human face.
        \\
        Text-to-Church & A normal church.
        \\
        Text-to-Cat & A cat.
        \\
        \bottomrule
    \end{tabular}
    \caption{Sentences we use to extract $CTE_{prompt}$s in different tasks.}
    \label{tab:TextPrompt}
\end{table}

\section{More Discussion on the Linear Operations in Cross-Modal Embedding Projection}
As we say in the main paper, the $CIE_{input}$ can be calculated from $CTE_{input}$ via the equation:
\begin{equation}
    CIE_{input} = CIE_{prompt} + (CTE_{input} - CTE_{prompt}).
    \label{equ:6}
\end{equation}
Why can we use such simple linear operations between the two latent spaces of CLIP? \citet{StyleGAN-nada, StyleCLIP} use similar linear assumption and have proved its correctness by practice. Here we discuss more details to prove it. 

The first way to illuminate it is to build a process of gradual manipulation. We assume there is a pair of semantically aligned text and image (\eg, a sentence ``A man with white hair'' and such an image). Then, we manipulate one single attribute of the image without changing any other attributes and edit the text with the same semantic (\eg, replace the white hair of the man in the image with black hair and change the sentence to ``A man with black hair''). The new text-image pair is still semantically aligned. So, the pair of their $CTE$ and $CIE$ still has high cosine similarity which will not be lower than the initial one. Because otherwise, we can keep doing such manipulations and the text-image pair will keep semantic alignment but their $CTE$ and $CIE$ will have low cosine similarity finally. In this case, it is a conflict with the character of CLIP. So, we have proved that when a change of one semantic occurs both in text and image, both of their $CTE$ and $CIE$ will change roughly collinearly.

The second perspective is to align the distance between $CTE_{input}$, $CTE_{prompt}$ and $CIE_{input}$, $CIE_{prompt}$. We assume a semantically matched pair of $CTE$ and $CIE$. Their distances to our prompt pair are $CTE - CTE_{prompt}$ and $CIE - CIE_{prompt}$. Due to the character of CLIP (embedding pair from semantically matched image-text pair has high cosine similarity), there will roughly be such an equation:
\begin{equation}
    CTE - CTE_{prompt} = CIE - CIE_{prompt}.
\end{equation}
From this, we can get Equation \ref{equ:6}. In practice, the equation usually be rewritten as Equation \ref{equ:7} with a constant scale factor $\alpha$ to control the distinctiveness.

\begin{table}
    \centering
    \begin{tabular}{p{1.3cm}p{0.6cm}p{2cm}p{1.2cm}p{1.2cm}}
        \toprule
        \ & Id & Block Name & In Size & Out Size
        \\
        \midrule
        \multirow{2}{*}{Head part} & 1-0 & FC+PReLU & 512 & 512
        \\
        & 1-1 & FC+PReLU & 512 & 512
        \\
        \midrule
        \multirow{15}{*}{Body part} & 2-0 & Dense block & 512 & 512
        \\
        & 2-1 & \$1-1 + \$2-0 & - & -
        \\
        & 2-2 & Dropout & - & -
        \\
        & 2-3 & Dense block & 512 & 512
        \\
        & 2-4 & \$2-2 + \$2-3 & - & -
        \\
        & 2-5 & Dropout & - & -
        \\
        & 2-6 & Dense block & 512 & 512
        \\
        & 2-7 & \$2-5 + \$2-6 & - & -
        \\
        & 2-8 & Dropout & - & -
        \\
        & 2-9 & Dense block & 512 & 512
        \\
        & 2-10 & \$2-8 + \$2-9 & - & -
        \\
        & 2-11 & Dropout & - & -
        \\
        & 2-12 & Dense block & 512 & 512
        \\
        & 2-13 & \$2-11 + \$2-12 & - & -
        \\
        & 2-14 & Dropout & - & -
        \\
        \midrule
        \multirow{2}{*}{Tail part} & 3-0 & FC+PReLU & 512 & 512
        \\
        & 3-1 & FC & 512 & 512
        \\
        \bottomrule
    \end{tabular}
    \caption{Specific architecture of \textit{C2S} projection network. ``\$'' denotes the layer output with the corresponding Id, ``FC'' denotes ``fully connected layer'', and all dropout layers have a dropout ratio of 0.1.}
    \label{tab:C2S_Arch}
\end{table}

\begin{table}[]
    \centering
    \begin{tabular}{p{0.6cm}p{2.5cm}p{1.5cm}p{1.5cm}}
        \toprule
        Id & Layer Name & In Size & Out Size
        \\
        \midrule
        0 & FC+BN+PReLU & 512 & 512
        \\
        1 & FC+BN+PReLU& 512 & 512
        \\
        2 & Concat(input, \$1) & - & 1024
        \\
        3 & FC+BN+PReLU & 1024 & 512
        \\
        4 & FC+BN+PReLU & 512 & 512
        \\
        5 & Concat(\$2, \$4) & - & 1536
        \\
        6 & FC+BN+PReLU & 1536 & 512
        \\
        7 & FC+BN+PReLU & 512 & 512
        \\
        8 & Concat(\$5, \$7) & - & 2048
        \\
        9 & FC+BN+PReLU & 2048 & 512
        \\
        10 & FC+BN+PReLU & 512 & 512
        \\
        11 & Concat(\$8, \$10) & - & 2560
        \\
        12 & FC+BN+PReLU & 2560 & 512
        \\
        13 & FC+BN+PReLU & 512 & 512
        \\
        \bottomrule
    \end{tabular}
    \caption{Specific architecture of each dense block. ``\$'' denotes the layer output with the corresponding Id, ``FC'' denotes ``fully connected layer'', and ``BN'' denotes ``BatchNorm layer''.}
    \label{tab:DenseBlock}
\end{table}

\begin{figure*}
    \centering
    \includegraphics[height=9.2cm]{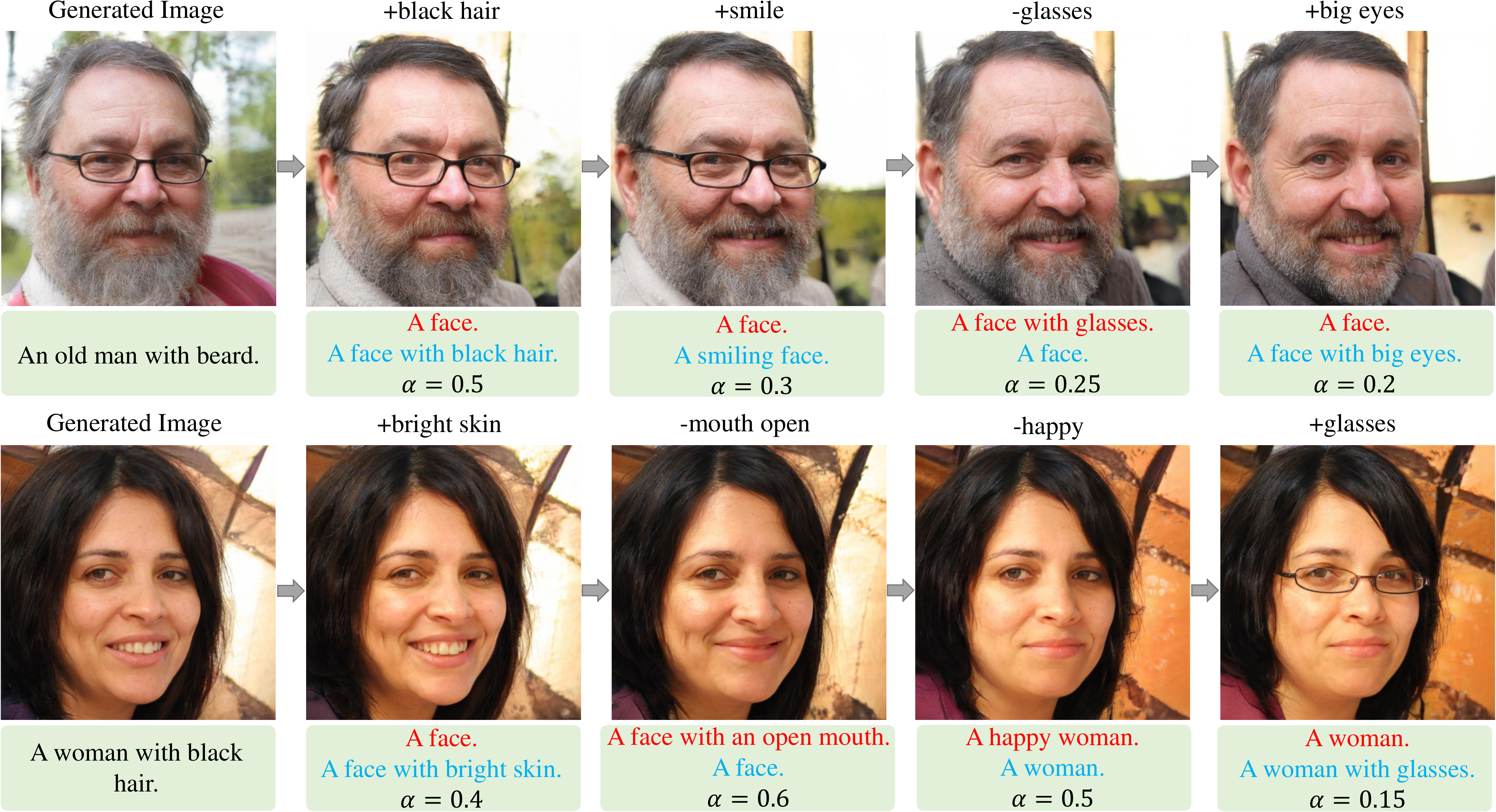}
    \caption{Manipulation results on generated images. We first generate realistic images and then manipulate them step by step. Every manipulation requires an \textcolor[RGB]{255, 0, 0}{origin text}, a \textcolor[RGB]{0, 176, 240}{target text} and a strength $\alpha$ (best view in color). ``+'' denotes adding certain attributes to the image and ``-'' denotes removing certain attributes from the image. The origin-target text pair is used to distinguish the attribute which the user wants to manipulate.}
    \label{fig:Manipulation}
\end{figure*}

\section{Manipulation on our Generated Images}
\label{sec:Manipulation}
In this section, we introduce a method of manipulating the images generated by our framework. This method helps the users finetune the generated images on those attributes which are not explicitly indicated in the input text description.

The manipulation method is similar to the generation method. The user should give a pair of texts which generally express the origin image (called origin text) and the target image (called target text) with a difference on the target attribute (\eg, if the user wants to lengthen the human's hair, she/he should give such a pair of sentences, ``A face.'', and ``A face with long hair.''). Our method extracts $CTE_{origin}$ and $CTE_{target}$ from them respectively. Then, we use a similar linear operation in our first module to calculate $CIE_{target}$. Taking $CIE_{origin}$ denote the $CIE$ of the originally generated image, we formulate this process as
\begin{equation}
    CIE_{target} = CIE_{origin} + \alpha \cdot (CTE_{target} - CTE_{origin}).
    \label{equ:7}
\end{equation}

It can be seen that this equation is similar to the projection from $CTE_{input}$ to $CIE_{input}$ in the first module. Then, via the second module, the $CIE_{target}$ is projected to the $SE$ from which we can generate realistic images. In practice, the constant $\alpha$ is usually set in the range of 0.05 to 0.7 to control the strength of the manipulation. 

In order to demonstrate the ability of our method of manipulation on generated images, we show some realistic results in Figure \ref{fig:Manipulation}. It should be noted that all the results in our main paper are not manipulated at all. This manipulation method we introduce here can give more freedom to our framework.

\begin{figure*}
    \centering
    \includegraphics[height=20cm]{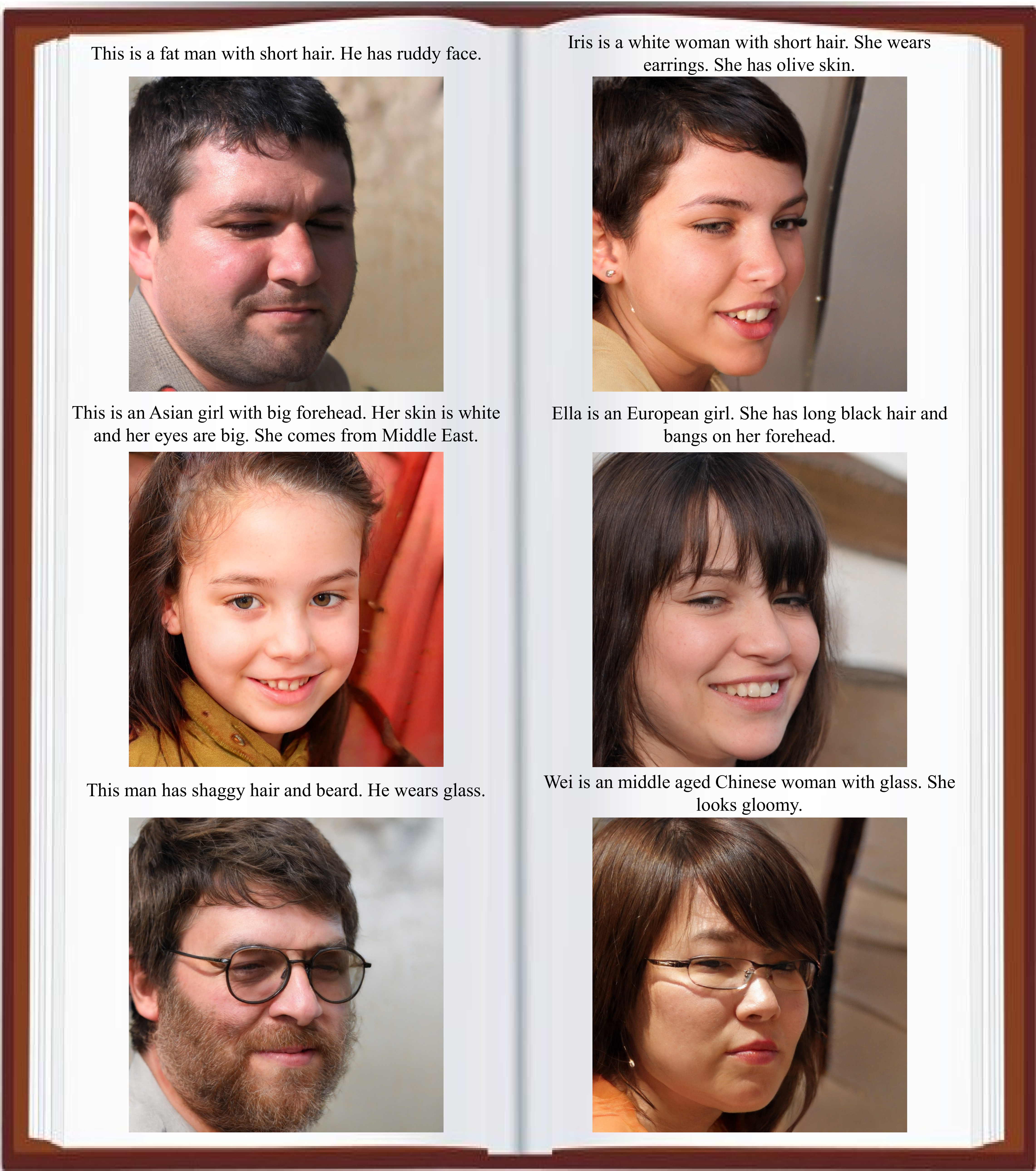}
    \caption{Additional realistic translation results of human faces of our framework. These results are not style transferred so that their attributes can be seen clearly without interference.}
    \label{fig:AdditionalResults1}
\end{figure*}

\begin{figure*}
    \centering
    \includegraphics[height=20cm]{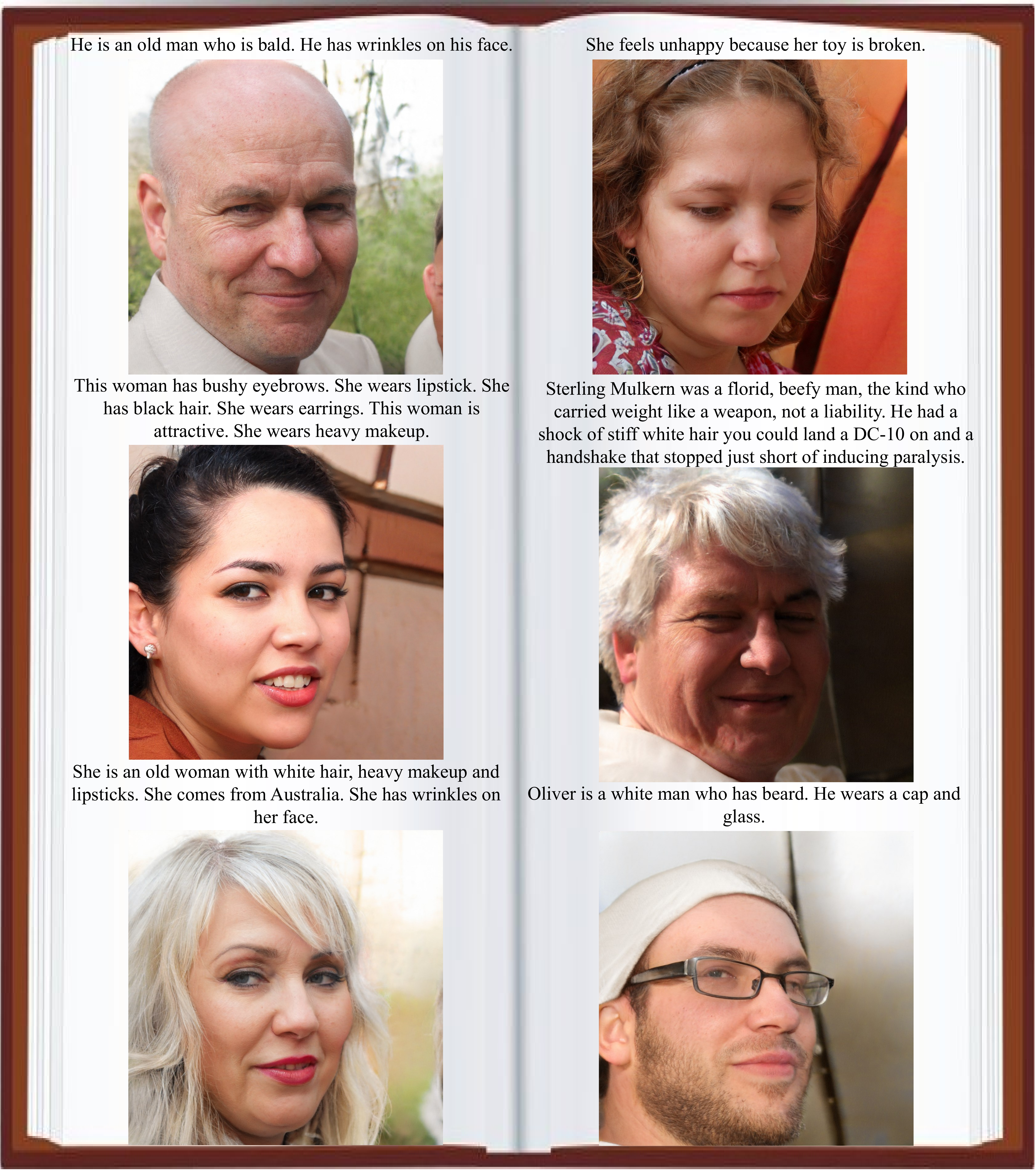}
    \caption{Additional realistic translation results of human faces of our framework. These results are not style transferred so that their attributes can be seen clearly without interference.}
    \label{fig:AdditionalResults2}
\end{figure*}

\begin{figure*}
    \centering
    \includegraphics[height=20cm]{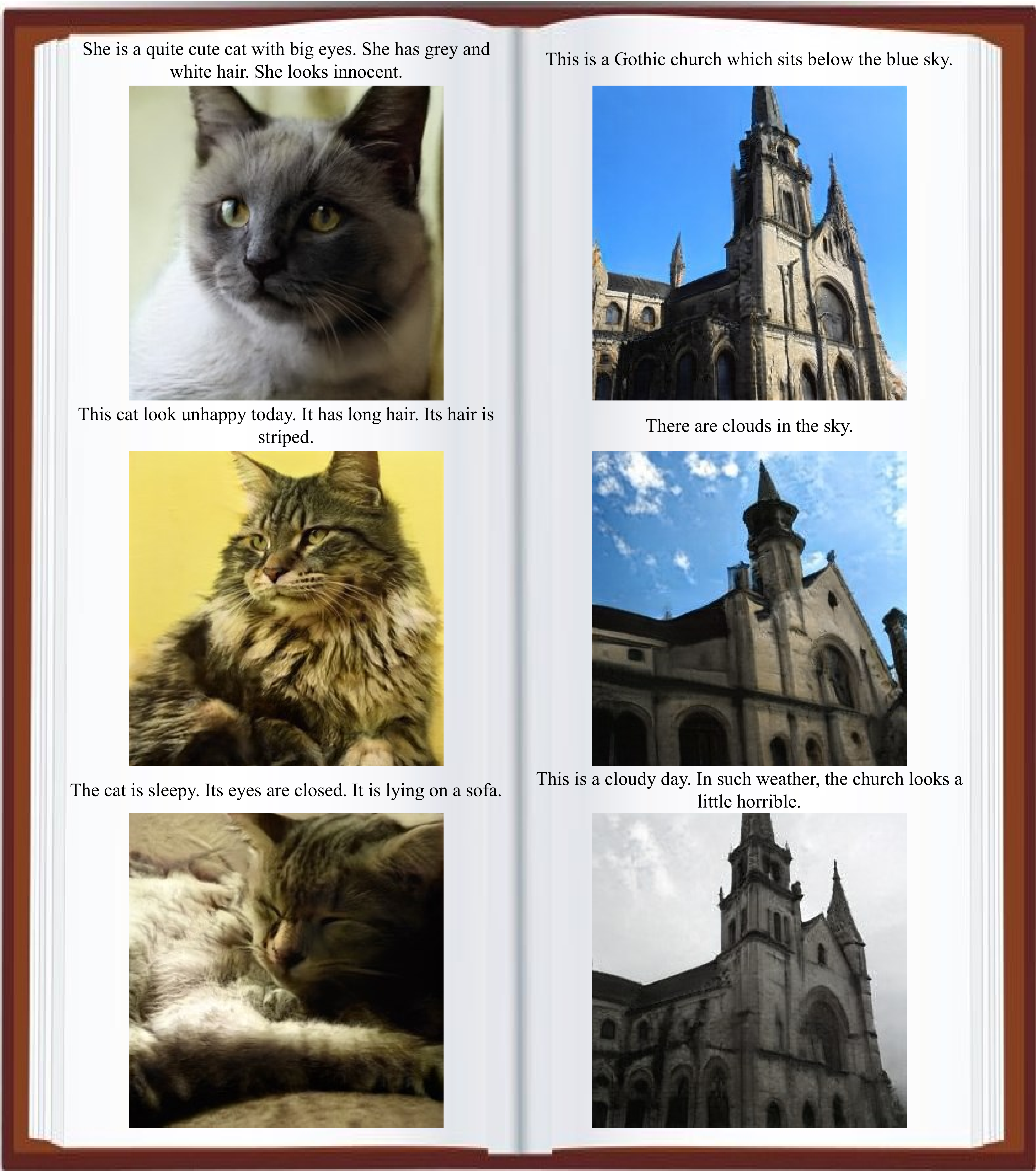}
    \caption{Additional realistic translation results of cats and churches of our framework. These results are not style transferred so that their attributes can be seen clearly without interference.}
    \label{fig:AdditionalResults3}
\end{figure*}

\begin{figure*}
    \centering
    \includegraphics[height=20cm]{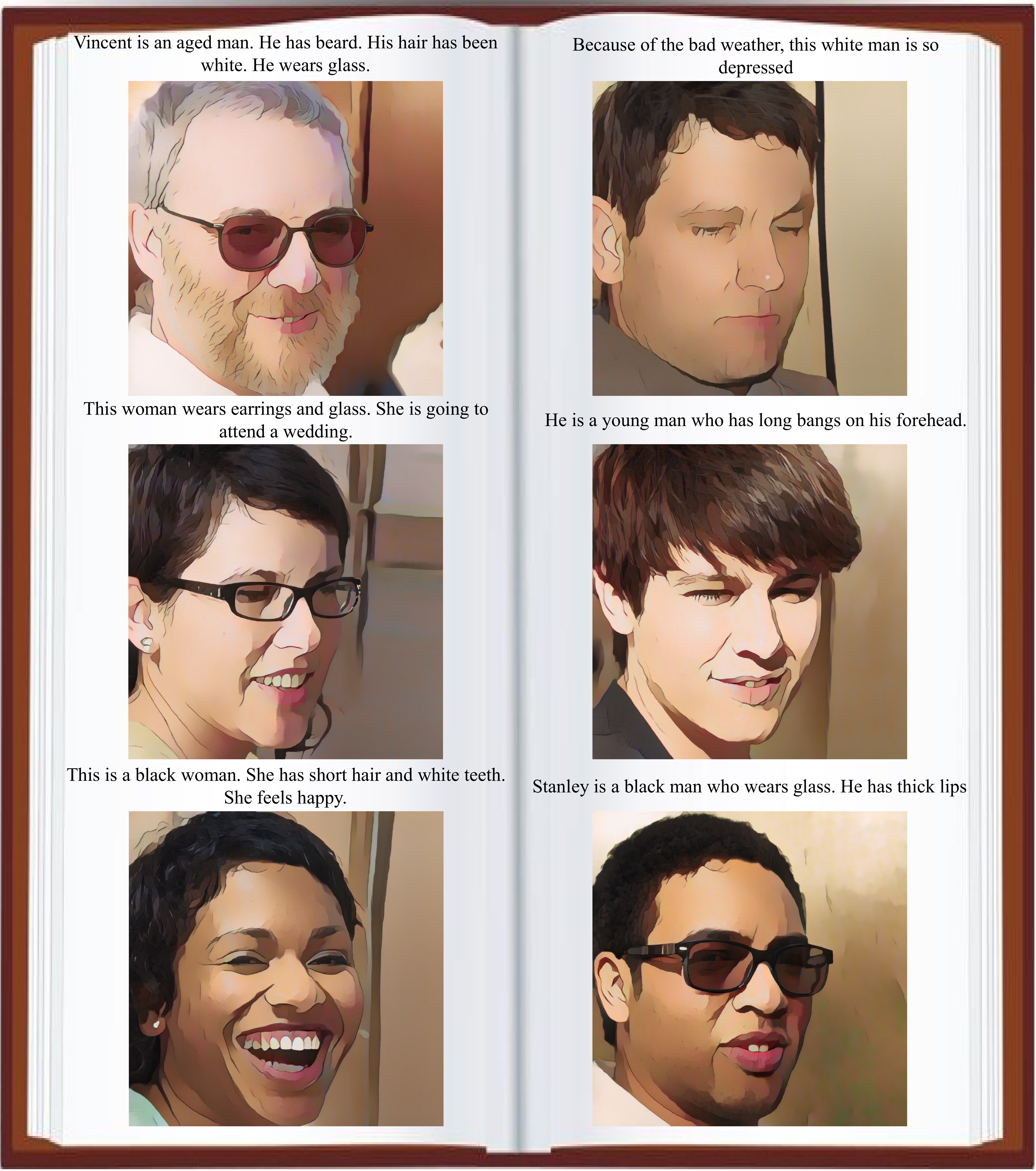}
    \caption{Additional illustration designs of human faces of our framework.}
    \label{fig:AdditionalResults4}
\end{figure*}

\begin{figure*}
    \centering
    \includegraphics[height=20cm]{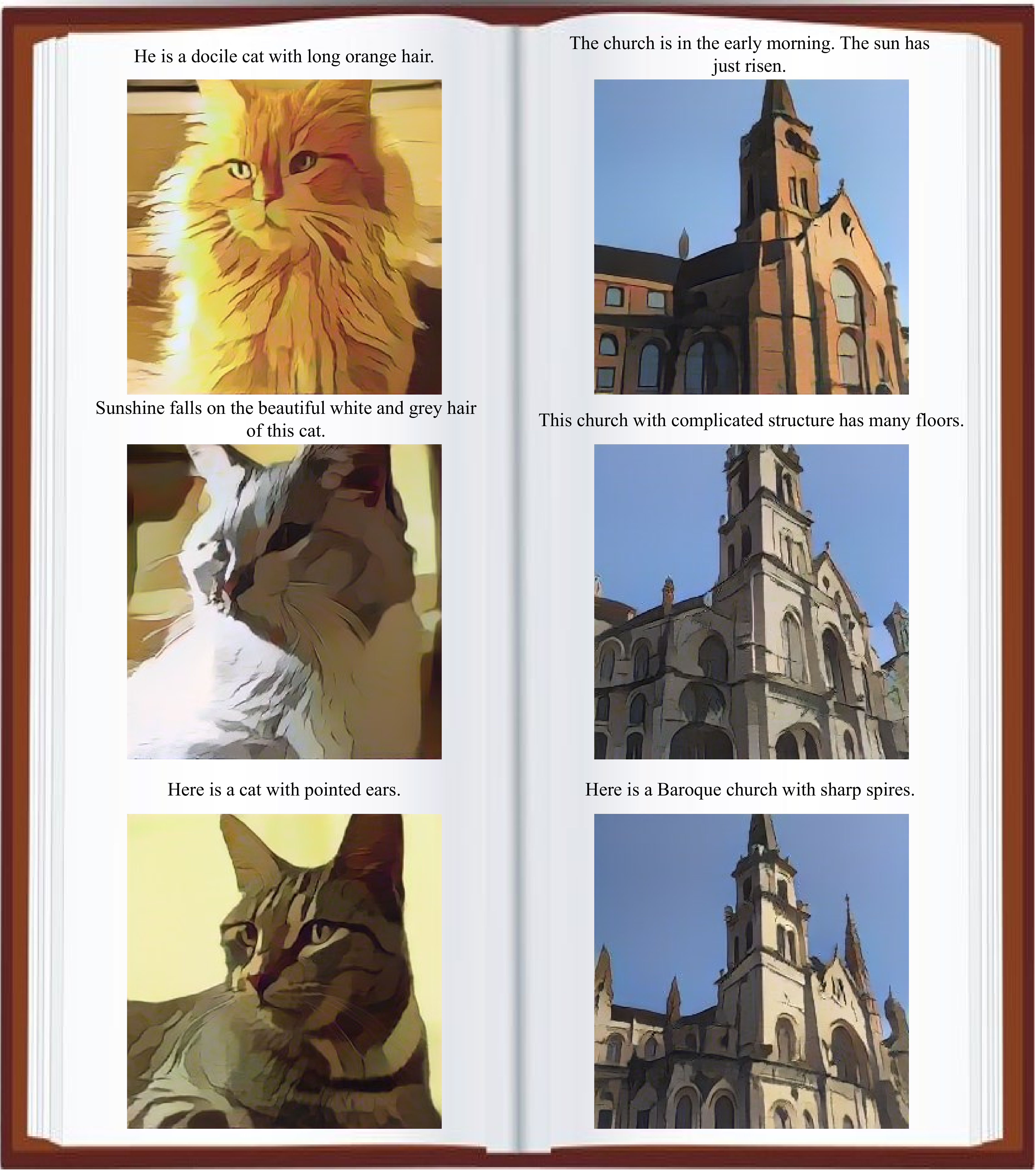}
    \caption{Additional illustration designs of cats and churches of our framework.}
    \label{fig:AdditionalResults5}
\end{figure*}

\section{Additional Samples}
\label{sec:Samples}
In this section, we show more realistic image results and illustration designs with high resolution to demonstrate the ability of our framework. All these images are generated without any random style mixing, so, they are completely reproducible. 

First, we show realistic images to clearly show the fine-grained attributes of the images we generate without the inference of style transfer. In Figure \ref{fig:AdditionalResults1}, \ref{fig:AdditionalResults2} and Figure \ref{fig:AdditionalResults3}, we show translation results of human faces and non-face cases respectively. 

Second, we show illustration designs to show the ability of our framework as an integral system. In Figure \ref{fig:AdditionalResults4} and Figure \ref{fig:AdditionalResults5}, we show illustrations of human faces and non-face cases respectively.

\clearpage

\end{document}